\documentclass{article}
\usepackage{ijcai26}
\usepackage{authblk}
\usepackage{times}
\usepackage{soul}
\usepackage{url}
\usepackage[hidelinks]{hyperref}
\usepackage[utf8]{inputenc}
\usepackage[small]{caption}
\usepackage{graphicx}
\usepackage{amsmath}
\usepackage{amsthm}
\usepackage{enumitem}
\usepackage{booktabs}
\usepackage[switch]{lineno}
\usepackage{twemojis}
\usepackage{graphicx}
\usepackage{amsfonts}
\usepackage{makecell}
\usepackage{xspace} 
\usepackage{booktabs}
\usepackage{multirow} 
\usepackage{graphicx}
\usepackage{colortbl}
\usepackage{xcolor}
\definecolor{lightgray}{gray}{0.9} 
\definecolor{lightgray2}{gray}{0.8} 
\usepackage[most]{tcolorbox}
\usepackage{amssymb}
\usepackage{soul}
\usepackage[table,dvipsnames]{xcolor}
\usepackage{booktabs}  
\usepackage{subcaption}
\usepackage[linesnumbered,ruled,vlined]{algorithm2e} 

\definecolor{mygreen}{RGB}{19,163,157}
\definecolor{myred}{RGB}{241,82,44}

\newenvironment{itemize*}%
 {\leftmargini=10pt\begin{itemize}%
  \setlength{\itemsep}{0pt}%
  \setlength{\parskip}{0pt}%
  }%
 {\end{itemize}}
\newenvironment{enumerate*}%
 {\leftmargini=10pt\begin{enumerate}%
  \setlength{\itemsep}{0pt}%
  \setlength{\parskip}{0pt}}%
 {\end{enumerate}}

\DeclareMathOperator*{\argmin}{arg\,min}

\definecolor{cyan10}{HTML}{E5F6FF}
\definecolor{cyan20}{HTML}{BAE6FF}
\definecolor{cyan60}{HTML}{0072c3}
\definecolor{cyan70}{HTML}{00539a}
\definecolor{cyan80}{HTML}{003a6d}
\definecolor{teal10}{HTML}{D9FBFB}
\definecolor{teal20}{HTML}{9EF0F0}
\definecolor{teal60}{HTML}{007d79}
\definecolor{orange10}{HTML}{FFF2E8}
\definecolor{orange20}{HTML}{FFD9BE}
\definecolor{orange60}{HTML}{ba4e00}
\definecolor{blue10}{HTML}{EDF5FF}
\definecolor{blue20}{HTML}{D0E2FF}
\definecolor{blue70}{HTML}{0043ce}
\definecolor{blue80}{HTML}{002d9c}
\definecolor{magenta10}{HTML}{FFF0F7}
\definecolor{magenta20}{HTML}{FFD6E8}
\definecolor{magenta30}{HTML}{ffafd2}
\definecolor{magenta50}{HTML}{ee5396}
\definecolor{magenta60}{HTML}{d02670}
\definecolor{magenta70}{HTML}{9f1853}
\definecolor{purple10}{HTML}{F6F2FF}
\definecolor{purple20}{HTML}{E8DAFF}
\definecolor{purple30}{HTML}{d4bbff}
\definecolor{purple70}{HTML}{8a3ffc}
\definecolor{rose10}{HTML}{FCF2ED}
\definecolor{rose20}{HTML}{F9D9D1}
\definecolor{rose60}{HTML}{ab5638}
\definecolor{rose70}{HTML}{853c27}
\definecolor{red10}{HTML}{FFF1F1}
\definecolor{red20}{HTML}{FFD7D9}
\definecolor{green10}{HTML}{DEFBE6}
\definecolor{green20}{HTML}{A7F0BA}
\definecolor{green70}{HTML}{0e6027}
\definecolor{green80}{HTML}{044317}
\definecolor{yellow10}{HTML}{fcf4d6}
\definecolor{yellow20}{HTML}{fddc69}
\definecolor{gray20}{HTML}{e0e0e0}
\definecolor{gray30}{HTML}{c6c6c6}
\definecolor{gray40}{HTML}{a8a8a8}
\definecolor{gray80}{HTML}{393939}

\newcommand{\ours}{\textsc{vera}\xspace}
\newcommand{\ie}{\textit{i.e.},\xspace} 
\newcommand{\eg}{\textit{e.g.},\xspace} 

\title{\scalebox{1.5}{\texttwemoji{eye}} \ours{}: Identifying and Leveraging Visual Evidence Retrieval Heads in Long-Context Understanding}

\author[1]{Rongcan Pei}
\author[2]{Huan Li}
\author[3]{Fang Guo}
\author[4]{Qi Zhu\thanks{Corresponding author.}\thanks{Work done outside Amazon.}}

\affil[1]{Tongji University, \texttt{prc@tongji.edu.cn}}
\affil[2]{Zhejiang University, \texttt{lihuancs@zju.edu.cn}}
\affil[3]{Westlake University, \texttt{guofang@westlake.edu.cn}}
\affil[4]{Amazon Web Services, \texttt{qzhuamzn@amazon.com}}

\begin{document}

\maketitle

\begin{abstract}
While Vision-Language Models (VLMs) have shown promise in textual understanding, they face significant challenges when handling long context and complex reasoning tasks. In this paper, we dissect the internal mechanisms governing long-context processing in VLMs to understand their performance bottlenecks. Through the lens of attention analysis, we identify specific \textbf{Visual Evidence Retrieval (VER) Heads} — a \textit{sparse, dynamic} set of attention heads critical for locating visual cues during reasoning, distinct from static OCR heads.
We demonstrate that these heads are causal to model performance; masking them leads to significant degradation. Leveraging this discovery, we propose \ours{} (Visual Evidence Retrieval Augmentation), a training-free framework that detects model uncertainty (\ie entropy) to trigger the explicit verbalization of visual evidence attended by VER heads. Comprehensive experiments demonstrate that \ours{} significantly improves long-context understanding of open-source VLMs: it yields an average relative improvement of \textbf{21.3\%} on Qwen3-VL-8B-Instruct and \textbf{20.1\%} on GLM-4.1V-Thinking across five benchmarks\footnote{The code is available at \url{https://github.com/Prongcan/VERA}.}. 

\end{abstract}

\section{Introduction}
Multi-modal documents, such as PDFs, academic papers, and financial reports, constitute the backbone of many modern information systems. Unlike plain text, these documents are composed of a complex interplay of text, layout, tables, and figures.
Early visual document processing pipelines rely on Optical Character Recognition (OCR) and subsequent natural language processing techniques. The emergence of Vision-Language Models (VLMs) has greatly eased the difficulty of end-to-end understanding on various components of multi-modal documents, enabling the unified interpretation of complex layouts. 

By treating document pages as high-resolution images, modern VLMs like GPT5.2 and Gemini3-Pro have achieved state-of-the-art on various visual and textual tasks. As agentic tasks demand increasingly long contexts, research has shifted toward investigating visual-text compression: determining whether visual modalities can represent textual content using significantly fewer tokens. To this end, DeepSeek-OCR~\cite{wei2025deepseek} sheds light on compressing 10$\times$ tokens on OCR tasks using a fine-tuned small VLM. 

\begin{figure}[t]
    \centering
    \includegraphics[width=1.0\columnwidth]{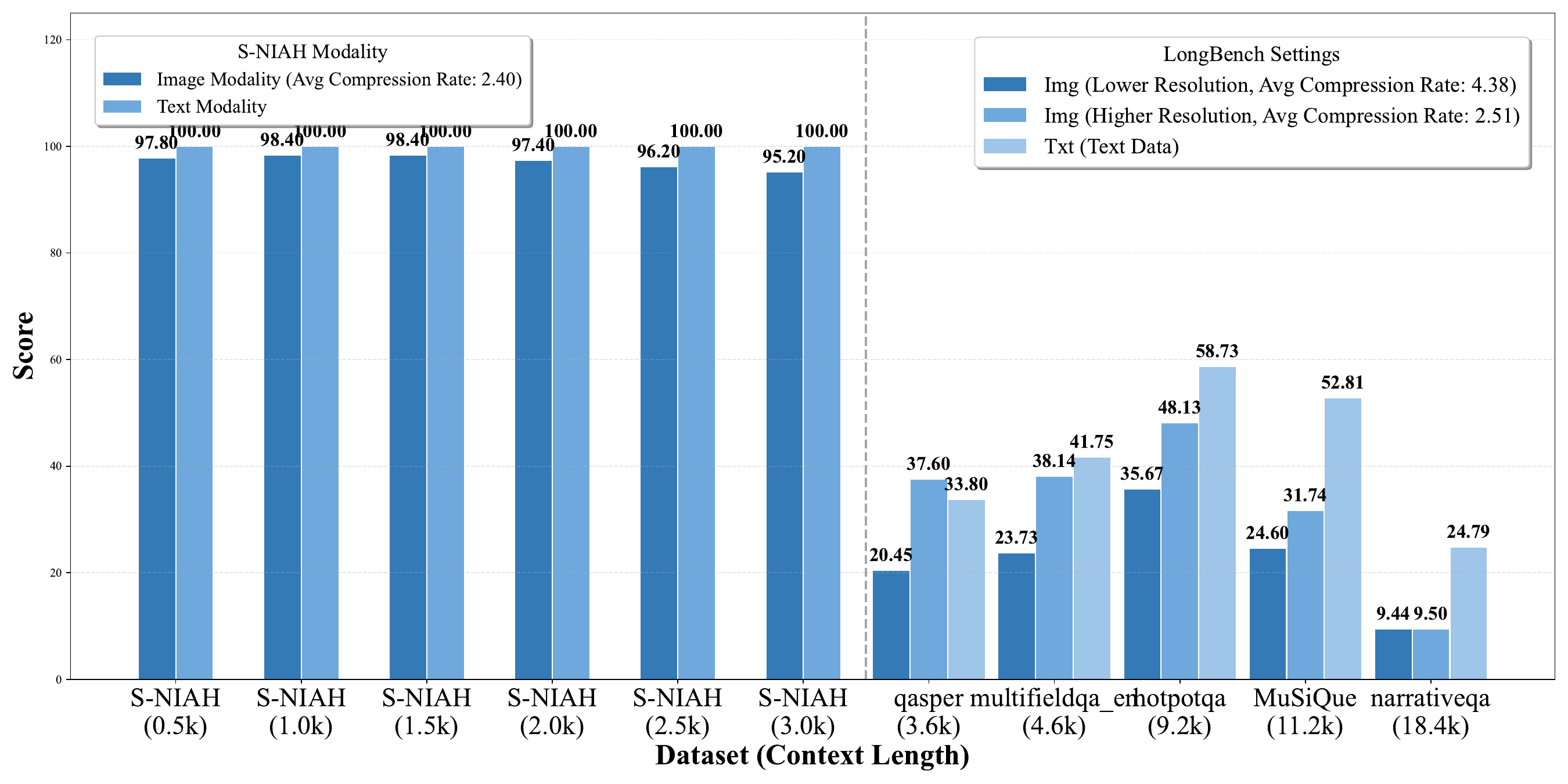} 
    \caption{Performance comparison between image and text inputs on the NIAH task (left) and long-context reasoning task (right).}
    \label{fig:longbench} %
\end{figure}

Recently, research on vision-text compression has been trending, while less attention has been paid to the conditions for achieving lossless compression.
We begin by evaluating the performance of Qwen3-VL-8B-Instruct on the ``Needle-in-a-Haystack'' task \cite{RULER}. We first convert the context into text-rich images, and then provide the input to the VLM in both raw text and image formats for a comprehensive comparison. We demonstrate that in various context lengths, Qwen3-VL-8B-Instruct consistently maintains a high accuracy level between 95.2\% and 98.4\%, as shown in the left part of Figure~\ref{fig:longbench}. These findings align with prior observations that VLMs can achieve performance parity with LLMs on simpler, lower-complexity tasks \cite{text_or_pixels}. 

However, this performance gap widens significantly as context length increases, particularly when evaluated on complex QA tasks (\eg multi-hop QA), as illustrated in the right panel of Figure \ref{fig:longbench}. On Narrative QA, the performance drop by switching input to image even exceeds 60\%!
In addition, we also observe a larger gap under lower resolution (higher compression ratio).
This observation reveals a critical bottleneck in VLM architectural capabilities that emerges as reasoning complexity and context length scale. We note that the selected resolution has already been optimized following the recommendations of recent work~\cite{Glyph}.
This gives rise to a question: \textit{what are the underlying internal mechanisms through which VLMs comprehend long-context textual information when it is presented as an image?}


Existing studies, such as CogDoc \cite{CogDoc} and Glyph \cite{Glyph}, started to enhance VLM's capability to comprehend text-rich images by designing end-to-end training frameworks and RAG methods; however, these approaches treat the model as a black box, yielding only marginal gains in our investigation. 

\begin{figure}[t]
    \centering
    \includegraphics[width=0.9\columnwidth]{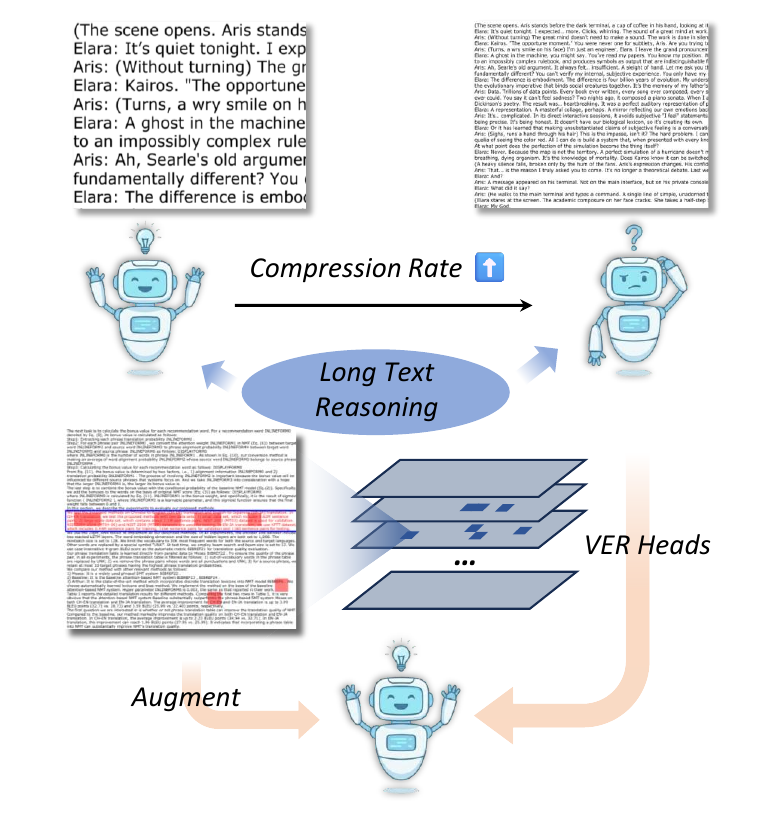} 
    \caption{Challenges in high-compression VLMs and illustration of visual retrieval augmentation.}
    \label{fig:teaser} %
\end{figure}

To bridge this gap, we perform in-depth attention analysis and identify a specific category named Visual Evidence Retrieval Heads (\textbf{VER Heads}) within VLMs, which are responsible for the precise localization of evidence regions within image patches. Through a comprehensive suite of controlled experiments (see Section~\ref{section3}), we confirm the universality and significance of these retrieval heads for long-context understanding. For reasoning models, we propose probing VER heads at the first high-entropy token. Our qualitative analysis suggests that this spike in uncertainty serves as a reliable proxy for the model's intrinsic ``retrieval moment,'' signaling the precise step where it actively seeks external visual evidence.

Motivated by this, in Figure~\ref{fig:teaser}, we devise an inference-time VLM augmentation framework \ours{} that transforms these mechanistic insights into actionable guidance. 
Instead of relying on external retrievers, \ours{} leverages the newly proposed VER heads to perform attention steering. Specifically, when a high-entropy state is detected, we extract the visual patches attended to by VER heads and explicitly verbalize them into textual context. In experiments, we find this insight-driven approach improves performance of both reasoning and non-reasoning VLMs on five different datasets by \textbf{20\%} on average. We believe the discovery of VER heads paves the way for high-ratio visual compression that preserves critical evidence.


We summarize our contributions as follows:
\begin{itemize}[nosep,leftmargin=*]
\item We formally defined Visual Evidence Retrieval (VER) heads and developed a quantitative metric to evaluate the alignment between attention maps and golden evidence, validating their existence across multiple datasets.

\item Through ablation studies, we established a causal link between VER head activation and task performance, providing a mechanistic explanation for the reasoning degradation observed in visual long-context understanding.

\item We introduced \ours{}, a training-free paradigm that improves performance by leveraging the model's underlying mechanistic logic without specialized training.
\end{itemize}

\section{Related Work}


\paragraph{Visual Document Understanding.}
Early approaches to multi-modal document processing relied on Optical Character Recognition (OCR)~\cite{ocr} to extract text, which was then processed using traditional NLP techniques. Large vision-language models like GPT-4o~\cite{hurst2024gpt} and Gemini2.5~\cite{comanici2025gemini} have since transformed this paradigm, enabling seamless joint perception and reasoning directly from visual document representations~\cite{ocrfree}. Recently, DeepSeek-OCR~\cite{wei2025deepseek} shed light on a third paradigm - utilizing VLMs for visual-text compression, thereby enabling the efficient processing of extremely long documents with significantly higher compression ratios. In this scenario, how to achieve high accuracy with low-resolution image input remains untouched.

\paragraph{Mechanistic Interpretability and Attention Analysis.}
Multi-head attention~\cite{vaswani2017attention} is the cornerstone of modern transformer-based language model architectures. Since the emergence of models like BERT and GPT-1/2/3, there has been extensive research~\cite{voita2019analyzing,clark2019does} into the specific functions of different attention heads, such as syntactic dependencies and coreference resolution. 
Subsequently, the field shifted toward identifying heads with distinct \textit{in-context} roles, most notably \textit{induction heads}~\cite{olsson2022context} discovered by Anthropic researchers. 
In the context of long-sequence modeling,~\cite{wu2024retrieval} identified \textit{retrieval heads} in LLMs—specific heads responsible for fetching information from distant text tokens. In the multi-modal domain, recent work has discovered OCR heads~\cite{baek2025large}, which are solely responsible for recognizing text within images. However, our work reveals a critical distinction: while OCR heads focus on low-level perception, they differ significantly from our proposed Visual Evidence Retrieval (VER) heads. Unlike the dense activation patterns of OCR heads, VER heads are sparse and functionally specialized—they are dynamically activated only during high-uncertainty ``retrieval moments'' to ground complex reasoning. To the best of our knowledge, \ours{} is the first work to identify and leverage these dynamic heads for long-context visual document understanding.

\paragraph{Long-context VLM Enhancement.}
In textual LLMs, dense retrieval was performed to identify the relevant chunks, while the paradigm shifted towards visual embedding-based retrieval for VLMs. For example, ColPali~\cite{copali} and PaliGemma~\cite{paligemma} leverage strong vision encoders to perform direct retrieval on image patches or page embeddings. 
More recently, research has explored solving long-document tasks end-to-end via adaptive visual compression mechanisms, such as Glyph~\cite{Glyph} and CogDoc~\cite{CogDoc}, though these approaches require extensive model post-training. Closely related to our work, Liu et al.~\cite{liu2025seeing} explore attention-guided inference-time intervention but are limited to a coarse-grained \textit{layer-wise} approach for standard VQA. In contrast, we propose \ours{}, a training-free framework that performs fine-grained \textit{head-wise} attention steering by identifying and explicitly verbalizing visual evidence.


\section{Characterizing VER Heads}
\label{section3}
\subsection{Definition and Identification Method}
\label{subsec:Definition and Identification}
Let $\mathcal{D} = \{(c_i, q_i, a_i, \mathcal{E}_i)\}_{i=1}^{N}$ denote a question answering dataset, where each sample consists of:
\begin{itemize}[nosep,leftmargin=*]
    \item A context passage $c_i$, constructed by concatenating multiple documents into a long context;
    \item A question $q_i$ and its corresponding answer $a_i$;
    \item A set of evidence spans $\mathcal{E}_i = \{e_1, e_2, \ldots, e_K\}$, where each $e_k$ is a contiguous text segment in $C$ that supports the answer.
\end{itemize}


\paragraph{Text-to-Image Rendering.}
Similar to Text-as-Pixels~\cite{text_or_pixels}, we render the context passage into an image $I = \texttt{Render}(c_i) \in \mathbb{R}^{H \times W \times 3}$. 
During rendering, we record the pixel-level bounding box $B_{e_k} = (x_{\min}, y_{\min}, x_{\max}, y_{\max})$ for each evidence span $e_k \in \mathcal{E}$.
The union of all evidence regions forms a binary mask $M \in \{0,1\}^{H \times W}$:
\begin{equation}
M_{x,y} = 
\begin{cases}
1 & \text{if } (x,y) \in \bigcup_{k=1}^{K} B_{e_k} \\
0 & \text{otherwise}
\end{cases}
\end{equation}
This method enables controlled evaluation of VLMs across varying context lengths and image resolutions.  

\paragraph{Visual Evidence Retrieval Score (VER Score).}
Given the rendered image $I$ and question $q_i$, we feed them into a VLM and extract the attention distribution when generating the most \textit{uncertain} token (\ie high entropy). 

The VLM's vision encoder divides the input image of size $H \times W$ into a grid of $H_{\text{patch}} \times W_{\text{patch}}$ patches, where $H_{\text{patch}}$ and $W_{\text{patch}}$ are determined by the encoder's patch size.
Each image patch $P_{i,j}$ corresponds to a visual token in the model's attention computation.
For each patch $P_{i,j}$, we denote $|P_{i,j}|$ as the number of pixels in patch $P_{i,j}$ and calculate evidence coverage weight
$w_{i,j} \in [0,1]$ as
\begin{equation}
w_{i,j} = \frac{1}{|P_{i,j}|} \sum_{(x,y) \in P_{i,j}} M_{x,y},
\end{equation}
measuring how much of patch $P_{i,j}$ overlaps with evidence regions.

Given an attention head $h$, let $A^{(h)}_{i,j}$ denote the attention weight assigned to the visual token corresponding to patch $P_{i,j}$ when generating the first output token.
The \textbf{patch-level visual retrieval score} is defined as
\begin{equation}
r^{(h)}_{i,j} = w_{i,j} \cdot A^{(h)}_{i,j}.
\end{equation}
The visual retrieval score of attention head $h$ is obtained by aggregating over all patches:
\begin{equation}
R^{(h)} = \sum_{i=1}^{H_{\text{patch}}} \sum_{j=1}^{W_{\text{patch}}} r^{(h)}_{i,j}.
\label{eq:visual_retrieval_score}
\end{equation}

 To quantify attention concentration independent of evidence size, we define the \textit{normalized visual retrieval score} as $\bar{R}^{(h)} = R^{(h)} / \rho$, where $\rho$ denotes the proportion of evidence pixels in the image. 
Finally, we identify the set of \textbf{Visual Evidence Retrieval (VER) Heads} as those satisfying $\bar{R}^{(h)} > \tau$, where the threshold $\tau$ is the midpoint of the normalized score range across all heads (i.e., $\tau = (\max \bar{R} + \min \bar{R}) / 2$).

\begin{figure}[h]
    \centering
    \includegraphics[width=1.0\columnwidth]{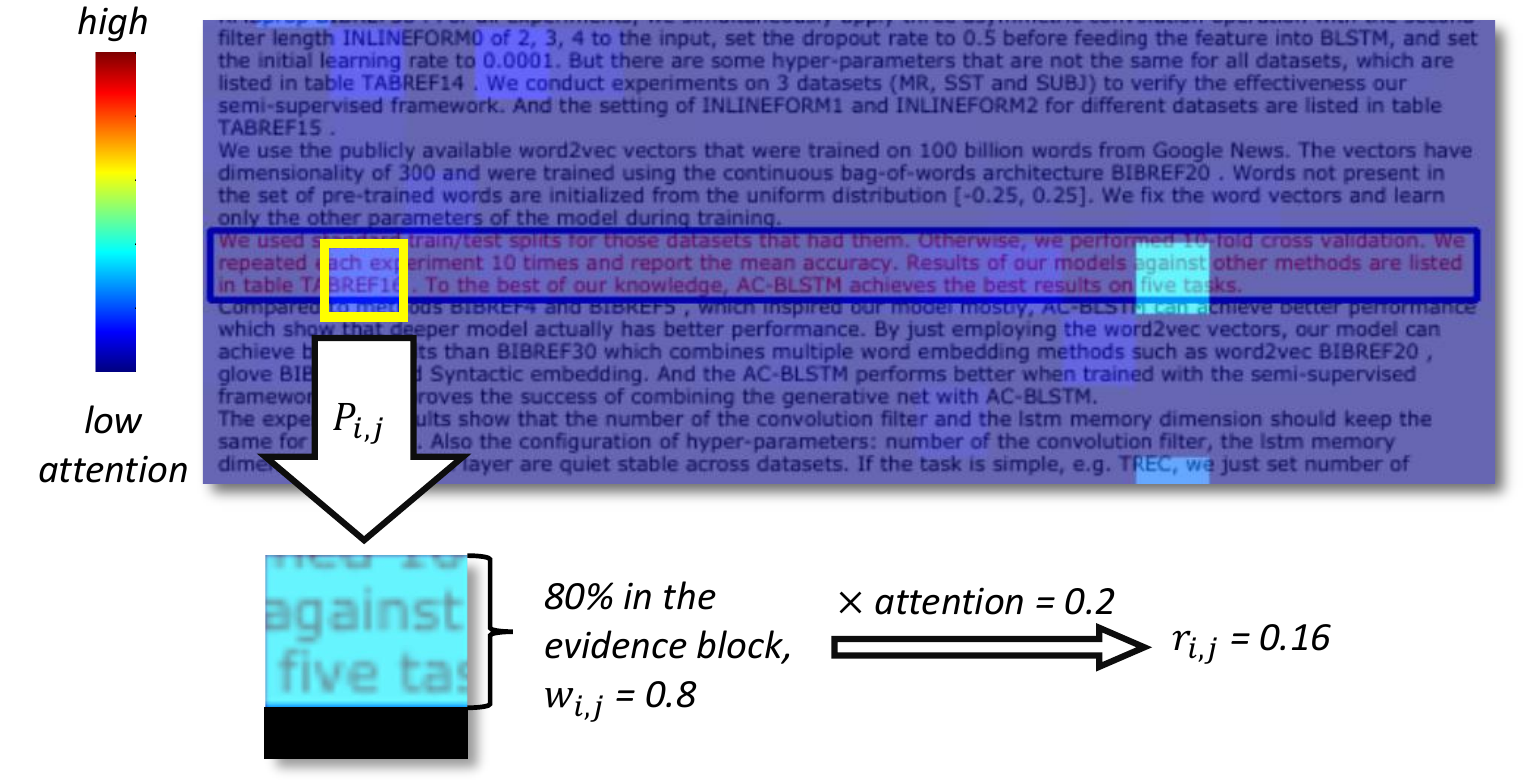} 
    \caption{Calculation of VER Score of a specific head.}
    \label{fig:compare} %
\end{figure}

\subsection{Quantitative Analysis of VER Heads}
\label{Quantitative Analysis of VER heads}

We now design experiments to study the effect of these VER heads across a diverse range of QA datasets and model architectures focusing on \textit{long-context reasoning}. The five different datasets are shown in Table~\ref{tab:datasets}.

We evaluate a range of popular VLMs, both reasoning and non-reasoning, covering different model families: 
instruction-following VLMs with different vision encoders and LLM backbone such as Qwen3-VL series, reasoning-enhanced VLMs including Qwen3-VL-8B-Thinking and GLM-4.1V-Thinking. 

\begin{table*}[th]
    \small
    \centering
    \caption{VER head Evaluation Results (F1 Score).}
    \begin{tabular}{lccccc}
        \toprule
         & NO mask (\%) & Random mask (\%)  & OCR head mask (\%) & VER head mask (\%) \\
        \midrule
        \multicolumn{5}{l}{\textbf{\textit{F1 Score of Qwen3-8B-VL on different datasets}}} \\
        DocMath & 3.47 & 4.24 & 4.62 & \textbf{2.80} \\
        Qasper & 28.37 & 26.63 & 26.75 & \textbf{18.81} \\
        HotpotQA & 28.48 & 26.60 & 28.27 & \textbf{23.80}\\ 
        Musique & 9.25 & 9.48 & 8.12 & \textbf{7.76}\\
        \midrule
        $\bar{\Delta}$ & 0 & -0.66 & -0.45 & \textbf{-4.10} \\
        \bottomrule
    \end{tabular}

    \label{tab:ver_mask}
\end{table*}

\begin{figure}
    \includegraphics[width=1.0\columnwidth]{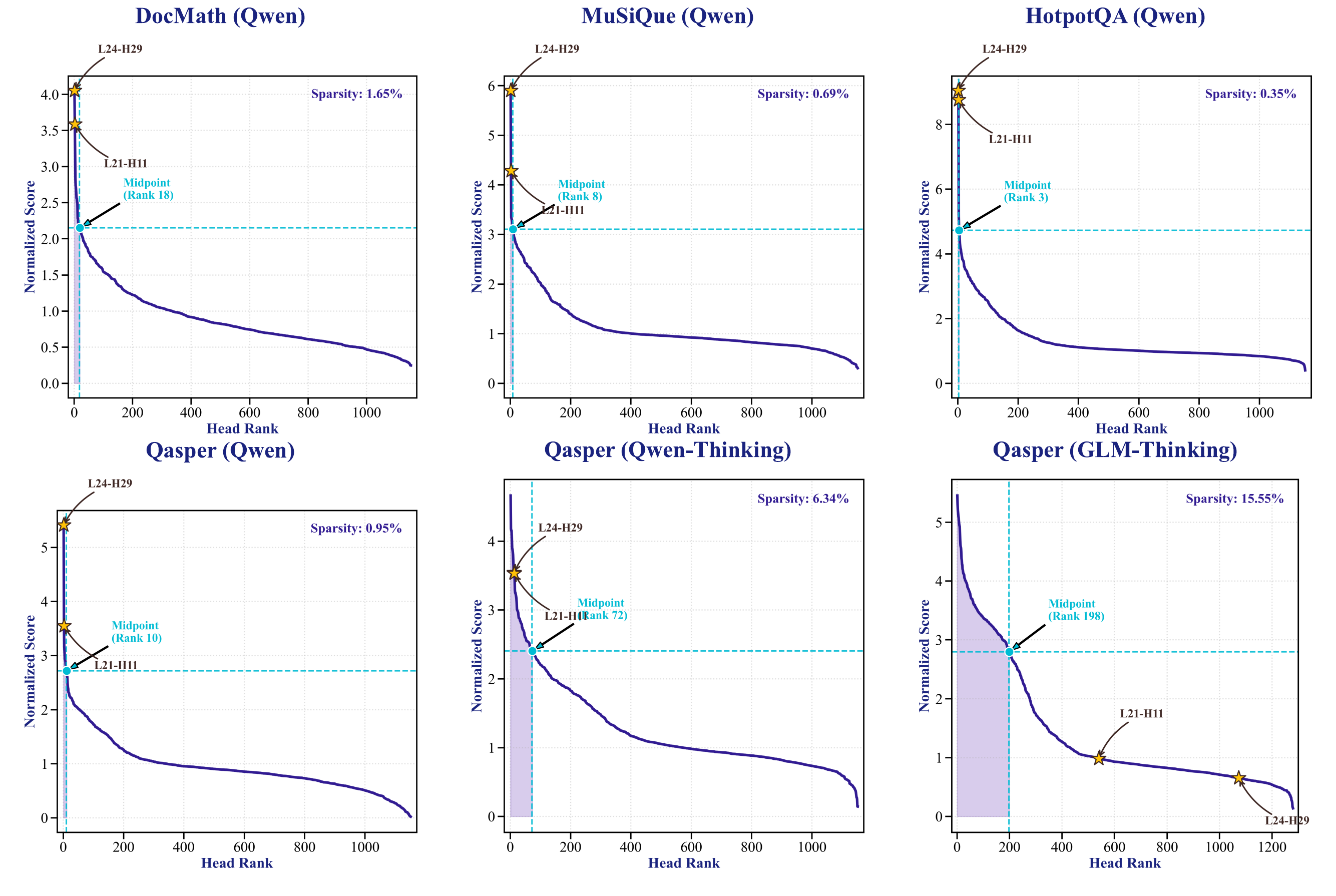} 
    \caption{The distribution of Visual Retrieval Scores across attention heads.}
    \label{fig:four_datasets_row} 
\end{figure}

To quantify VER head activation, we compute the Visual Retrieval Score (defined in Section~\ref{subsec:Definition and Identification}) for each of the $L \times H$ attention heads across all datasets. 
To establish causality, we further conduct ablation experiments by masking the top-5 heads of varying types (VER, OCR, and random) and measure the resulting performance changes.
Results are presented in Table~\ref{tab:ver_mask}. Throughout comprehensive experiments, we have the following observations:

\paragraph{(1) VER heads are \textit{sparse} yet \textit{universal}—they exist across different VLMs and remain consistent across datasets.}
In Figure~\ref{fig:four_datasets_row}, we plot the average visual retrieval score of all heads on different datasets and models. Across all evaluated datasets, only a small fraction ($<$1.65\%) of attention heads of Qwen3-8B-VL exhibit high visual retrieval scores and are defined as VER heads. This explains the performance degradation of VLMs when used for long-context tasks. In the same model, we often observe a fixed set of VER heads are frequently activated in different datasets. For example, head (24,29) and (21,11) of Qwen3-VL-8B demonstrate high VER scores in all tested datasets. Finally, we examine the VER head distribution in reasoning-enhanced models, such as Qwen-Thinking and GLM-Thinking. As illustrated in Figure~\ref{fig:four_datasets_row}, these models exhibit a similar sparse, long-tail distribution. Notably, the heightened activation intensity observed in these heads directly correlates with the superior performance recorded on downstream task metrics.
\begin{figure}
    \centering
    \includegraphics[width=1.0\columnwidth]{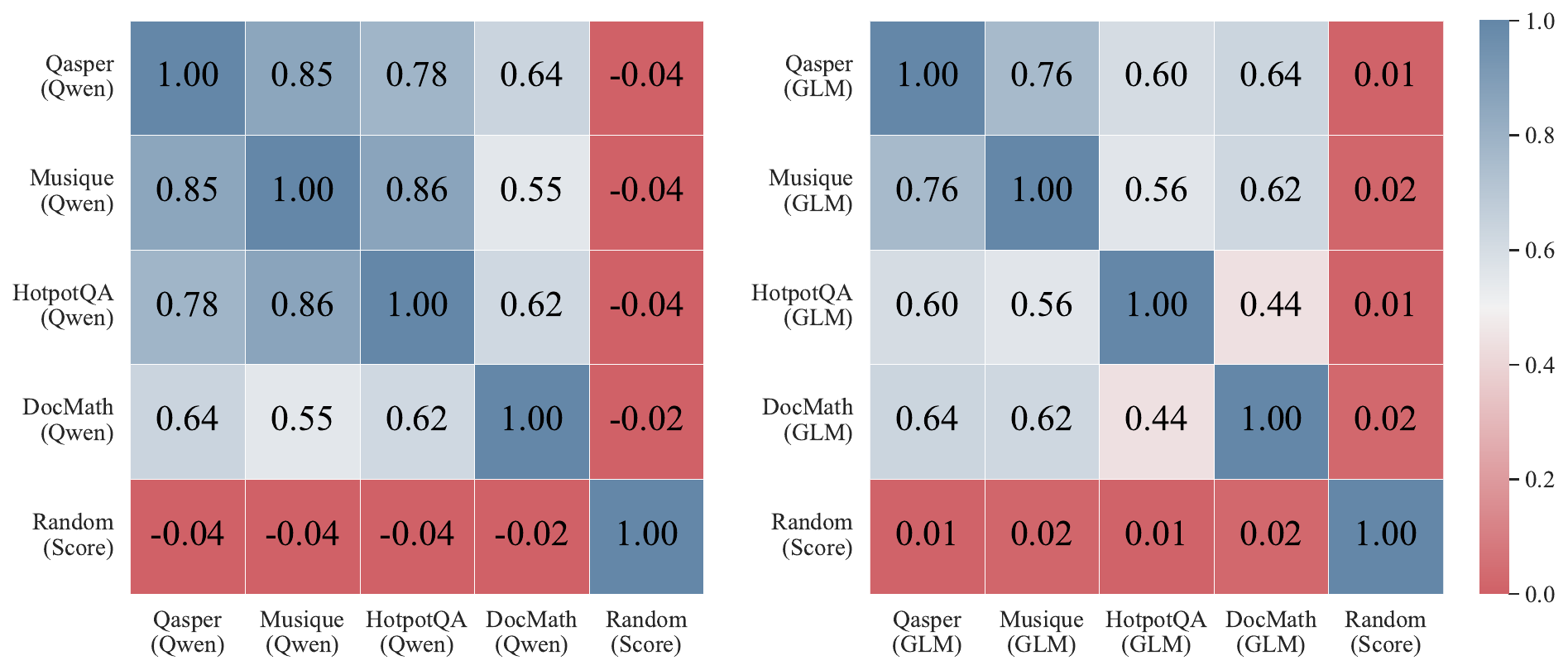} 
    \caption{Spearman's rank correlation coefficients of Visual Retrieval Score distributions across different datasets.}
    \label{fig:spearman} %
\end{figure}

To quantify this consistency in different datasets, we compute the Spearman rank correlation coefficient between the vectors of flattened attention head's visual retrieval score of paired datasets. 
As illustrated in Figure \ref{fig:spearman}, the Spearman rank correlation coefficients between the attention head rankings of diverse datasets are remarkably high, consistently exceeding \textbf{0.44}. This indicates a strong functional alignment, where the model consistently prioritizes the same specific set of heads for visual retrieval tasks regardless of the data distribution.

\paragraph{(2) VER heads are \textit{functionally critical} for long-context reasoning—masking them degrades performance, while masking other head types does not.
}
In Table~\ref{tab:ver_mask}, we investigate the performance change of Qwen3-8B-VL by masking most frequent  (\ie top-5) VER heads, OCR heads~\cite{baek2025large} and random heads. The result shows  masking OCR Heads does not lead to substantial performance degradation on the evaluated datasets; on average, OCR heads do not perform statistically differently from masking the same amount of random heads. Masking VER heads causes F1 Scores drop by 4.10\% on average, compared with a drop by 0.66\% after masking OCR heads and 0.45\% after masking random heads, which validates their importance in long-context understanding. 

\begin{figure}
    \centering
    \includegraphics[width=1.0\columnwidth]{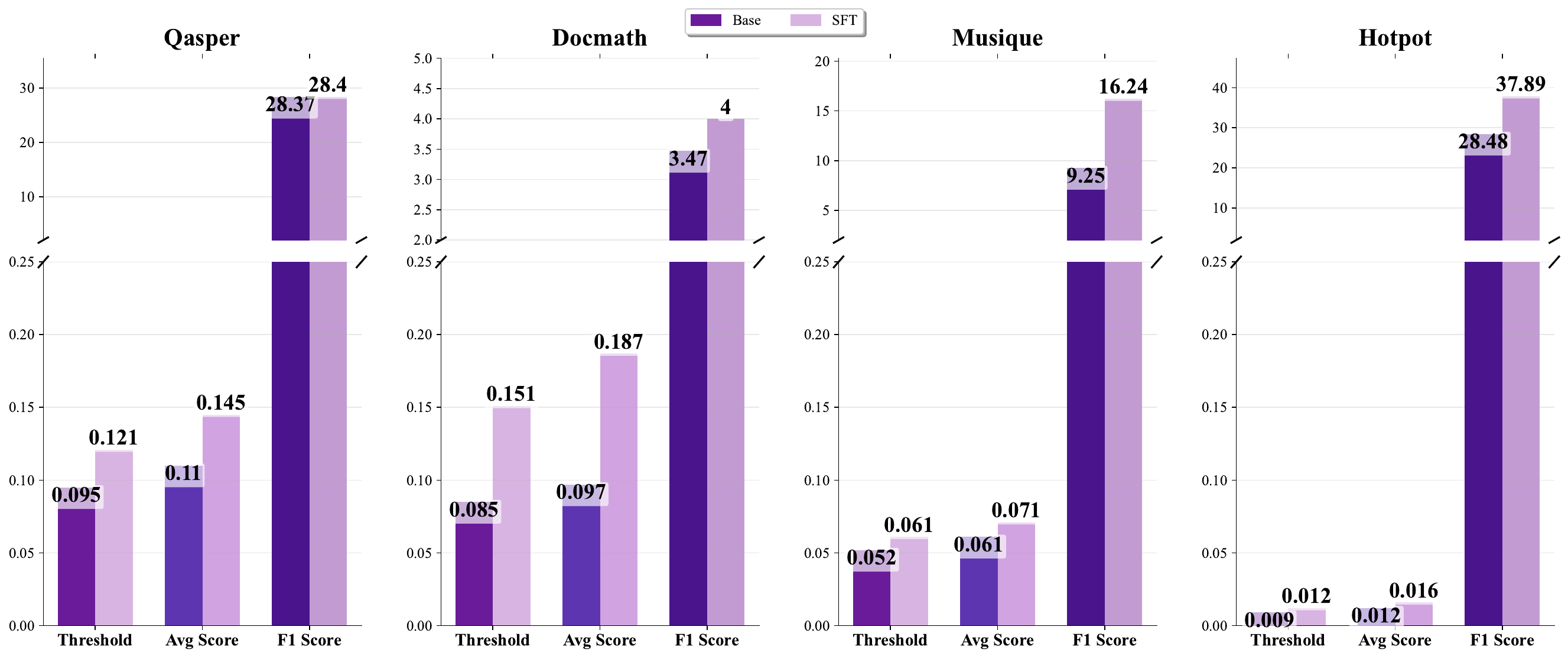} 
    \caption{The threshold demarcating the top 5\% of scores, average VER Scores of the top 5\% heads, and dataset performance before and after fine-tuning.}
    \label{fig:SFT_circle} %
\end{figure}

\paragraph{(3) Does fine-tuning activate new VER heads, enhance existing ones, or both?}
Having established that VER heads are functionally critical for long-context reasoning, a natural question arises regarding their plasticity: how do these specific mechanisms evolve during training? We investigate whether the performance gains observed after supervised fine-tuning are mechanistically rooted in the intensification of existing VER heads or the emergence of new ones. We first reorganized the Hotpot training dataset into a format consisting of image context, textual questions, and ground-truth answers, totaling 45,000 samples. Subsequently, we performed full-parameter supervised fine-tuning on the Qwen3-VL-8b-Instruct model.
In Figure~\ref{fig:SFT_circle}, we found that with the improvement of performance, the activation of VER Head increased significantly across multiple datasets following SFT. As shown in Figure~\ref{fig:SFT_circle}, on the Hotpot dataset, the average score of the heads ranked in the top 5\% by visual retrieval score increased by approximately 25\% after fine-tuning. On the DocMath dataset, this rate of increase reached 90\%, demonstrating strong generalizability.
We further compared Visual Retrieval Score distributions before and after SFT, confirming a strong correlation between improved comprehension and increased VER head activation.

\begin{figure*}[t] 
    \centering
    \includegraphics[width=0.8\textwidth]{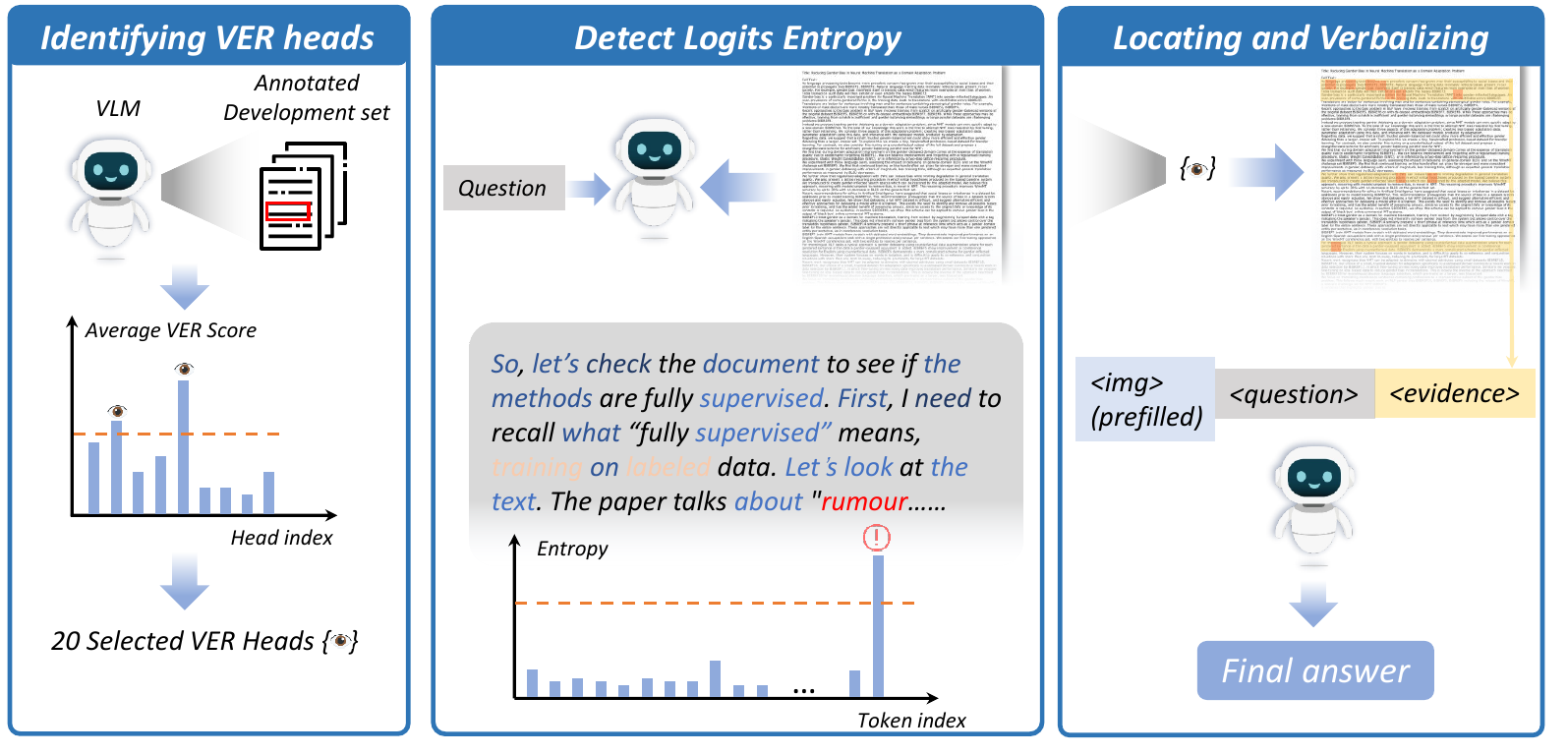} 
    \caption{Pipeline of \ours{}.}
    \label{fig:pipeline}
\end{figure*}

\section{Harnessing VER Heads in VLM Reasoning}
\label{pipeline}
In Section 3, we carefully examine the causal relations between the activation of VER Heads and the model's ability to reason over long contexts. 
This finding suggests that VLMs inherently possess the mechanism to locate task-specific visual information, but this capability is often diluted, which is also noted in recent work~\cite{spatial_reason} as imbalance visual and textual attention. 

Building on this insight, we propose \ours{} (Visual Evidence Retrieval Augmentation), a training-free method that steers model focus by converting patches with high VER-head activation into text, effective for both instruction-following and reasoning VLMs. The complete pipeline of \ours{} is illustrated in Figure \ref{fig:pipeline}.

\subsection{Identifying VER Heads for Different Models}
In instruction following VLMs, we observe the model will attempt to retrieve the relevant visual information at the very first token, which makes VER head identification straightforward following Equation.

However, the attention distribution of reasoning models at the initial decoding step often fluctuates. Consequently, the first token fails to reliably attend on evidence regions $M$. To accurately characterize VER heads in these models, we must instead identify specific ``retrieval moments'' where the model actively seeks visual information to answer the question.

\paragraph{Logits Entropy as a Reasoning Indicator.} 
We premise our approach on the observation that `retrieval' is a consequence of `uncertainty'. Specifically, we utilize logits entropy, a widely adopted metric for quantifying model uncertainty~\cite{logits-entropy}.
Let $\mathcal{H}_t$ denote the entropy of the probability distribution over the vocabulary at generation step $t$. Qualitative analysis reveals that tokens with high $\mathcal{H}_t$ typically correspond to pivotal steps in the chain-of-thought where the model explicitly seeks external visual information. 
\paragraph{The First High-Entropy Token.} Based on this motivation, we propose identifying VER heads by analyzing the attention distribution $A^{(h)}$ at the first high-entropy token. We define this target step $t^*$ as:\begin{equation}
t^* = \argmin \{ t \mid \mathcal{H}_t > \delta \}
\end{equation}where $\delta$ is a hyper-parameter. In our experiment, we set $\delta = 2.0$ to effectively isolate these ``retrieval moment'' from standard generation steps, where logits entropy is typically negligible ($10^{-1}\sim 10^{-7}$). We select the first high-entropy token for two primary reasons: (1) indicating uncertainty on visual context; (2) targeting the initial instance minimizes interference from generated reasoning history, ensuring the attention signal reflects a reliance on the source image $I$ rather than textual cues from the chain-of-thought.
Upon identifying $t^*$, we compute the visual retrieval score for head $(i,j)$ $R^{(h_{ij})}$ regarding Equation~\ref{eq:visual_retrieval_score}.
\subsection{Locating Visual Evidence Patch via Head Selection}
To accurately locate visual evidence during inference for a target model, we must first identify a specific set of VER heads that reliably attend to evidence regions across different datasets. Building on our observation in Section~\ref{section3} 
that VER heads are highly similar in different datasets, we employ a straightforward selection strategy based on a held-out dataset with ground-truth annotations.

\paragraph{Head Selection Strategy.} We construct a development set $D_{\text{dev}}$ consisting of a small number of annotated instances.
To determine the utility of each attention head, we compute the average visual retrieval score for all attention heads. Then we select the set of active retrieval heads, $H_\text{selected} = \{h_1, ..., h_k\}$, as the top-$k$ heads with the highest average scores on $\mathcal{D}_{dev}$.

\paragraph{Evidence Patch Extraction.}
During inference time, we average the attention maps of the selected heads in $H_\text{selected}$.  We then extract the top-$N$ visual patches as evidence patches $\mathcal{E} = \{P_{i,j}\}$ with the highest accumulated attention weights. These patches serve as the ``visual evidence" for the subsequent verbalization step. 

\subsection{Verbalizing Retrieved Visual Evidence}
To explicitly steer the model's reasoning, we construct a retrieval-augmented prompt that integrates the visual cues. Formally, the input sequence $\mathbf{X}$ is structured as:
\begin{align}
\label{eq:verbalization}
\mathbf{X} = [\texttt{<image>}, \mathbf{C}_{ver}, \texttt{Question}, \texttt{Answer:}]
\end{align}
where $\texttt{<image>}$ represents the visual tokens and $\mathbf{C}_{ver}$ denotes the verbalized text derived from the visual evidence patches $\mathcal{E}$.

To mitigate semantic fragmentation, we employ a horizontal context expansion strategy: rather than verbalizing isolated patches, we extract the entire text row corresponding to the vertical coordinates of any selected patch $P \in \mathcal{E}$. This ensures that the retrieved evidence maintains local textual consistency. While this verbalization can generally be achieved via OCR models, in our experiments—where images are rendered from source text—we directly map the retrieved visual coordinates back to the corresponding source text lines for precise evaluation. 
Inference of \ours{} can be optimized via caching the KV of image token in first pass. And the detailed algorithm is shown in Appendix A.

\begin{table*}[h]
\centering
    \caption{Performance comparison (F1 \%) of different RAG methods across long-context QA benchmarks. VER-RAG consistently outperforms baselines by leveraging attention from identified VER heads. The best performing methods in each column are highlighted in \textbf{bold}, and the second-best are \underline{underlined}.}
    \label{tab:ver-rag}
\resizebox{\textwidth}{!}{
\begin{tabular}{l c ccc cc c | c} 
    \toprule
    \multirow{2}{*}{Method} & 
    \multirow{2}{*}{DocMath} & 
    \multicolumn{3}{c}{Qasper} & 
    \multirow{2}{*}{HotpotQA} & 
    \multirow{2}{*}{MuSiQue} & 
    \multirow{2}{*}{\shortstack{LongBench\\Pro}} &
    \multirow{2}{*}{Rel.$\Delta$}\\
    \cmidrule(lr){3-5}
    & & Extract & Abstract & Bool & & & \\ 
    \midrule
    \multicolumn{8}{l}{\textit{Baselines}} \\
    Qwen3-VL-8B-Instruct          & 3.47  & 25.54 &17.82 &70.36 & 28.48 & 9.25  &  27.56 & / \\
    Qwen3-VL-8B-Instruct Random RAG      & 5.61  & 23.96 & 16.24 & 65.86 & 30.16 & 14.68 &  28.00 & 1.1\%\\
    Qwen3-VL-8B-Instruct OCR RAG         & 4.72  & 25.15 & 17.99 & 71.48 & 29.03 & 14.70 & 26.40 & 3.8\% \\
    GLM-4.1V-9B-Thinking & 15.71 & 41.69 & 26.71 & \underline{81.28} & 39.49 & 27.85 & \underline{31.06} & / \\
    GLM-4.1V-9B-Thinking Random RAG & \underline{21.99} & 46.92 & \underline{29.61} & 80.47 & 38.00 & 28.71 & 28.84 & 4.1\% \\
    GLM-4.1V-9B-Thinking Embedding RAG & 17.49 & 45.21 & 23.71 & 73.60 & 36.03 & 28.08 & 26.29 & -5.1\% \\
    GLM-4.1V-9B-Thinking ColPali RAG & 21.18 & \underline{47.51} & 26.74 & 75.16 & \textbf{43.89} & \underline{30.24} & 28.58 & 3.6\%\\
    Glyph           & 13.61 & 39.26 & 24.45 & 45.62 & 38.74 & 24.87 & 28.94 & /\\
    \midrule
    \textit{Attention-Guided RAG} &  & \\
    \textbf{Qwen3-VL-8B-Instruct \ours{} (Ours)} & 9.45 & 39.29 & 22.56 & 71.12 & 32.32 & 17.84 & 28.74 & 21.3\%\\
    \textbf{GLM-4.1V-9B-Thinking \ours{} (Ours)} & \textbf{29.02}& \textbf{54.57} & \textbf{34.67} & \textbf{94.23} & \underline{39.56} & \textbf{30.58} & \textbf{34.20} & 20.1\%\\  
    \bottomrule
\end{tabular}
}
\end{table*}

\section{Experiments}

To evaluate the effectiveness of \ours{} in enhancing VLM's ablility in long-context understanding comprehensively, we applied it to both instruction-following and reasoning models. Through extensive experiments across multiple datasets. We aim to answer the following research questions:

\textbf{RQ1: How does \ours{} perform compared to existing baselines?} 

\textbf{RQ2: How well does \ours{} generalize across different models and datasets?} 

\textbf{RQ3: Is \ours{} sensitive to hyper-parameters?}

\textbf{RQ4: Is the performance of \ours{} positively correlated with its retrieval capability?}

\subsection{Evaluation Setup}
\label{setup}
To validate the effectiveness of the \ours{} framework, we focus on evaluating VLMs in long-context Question Answering (QA) task. Following the methodology introduced in Section~\ref{subsec:Definition and Identification}, we transform standard text-based QA benchmarks into visual documents. This rendering process is optimized to achieve high information density, allowing us to simulate long-context visual understanding tasks under the same controlled experimental environment.


\paragraph{Datasets.} 
We select five different datasets\cite{docmath,qasper,hotpotqa,musique,longbenchpro} as shown in Table~\ref{tab:datasets} based on three criteria: (1) \textbf{diverse domains}: covering mathematical reasoning, scientific literature, and multi-hop QA\footnote{We augment HotpotQA and Musique into long-context datasets, and clean the LongBench Pro with details in Appendix B.} to ensure generalizability; (2) \textbf{annotated evidence}: some datasets contain ground-truth evidence annotations; (3) \textbf{different question complexity}: featuring QA pairs with varying difficulty levels (\eg extractive, abstractive). Specifically, we sample 70 instances from each seen dataset to construct $\mathcal{D}_\text{dev}$ for model-specific VER heads detection. 
\begin{table}[h]
    \centering
    \small 
    \caption{Overview of datasets used in our experiments.}
    \setlength{\tabcolsep}{4pt} 
    \begin{tabular}{lccc} 
        \toprule
        Dataset & \makecell{Context\\Length} & \makecell{Avg. visual \\ tokens} & Seen in $\mathcal{D}_\text{dev}$ \\ 
        \midrule
        DocMath             & 2.2k & 1.3k & Yes \\
        Qasper              & 3.3k & 1.5k & Yes \\
        HotpotQA  & 9.5k & 5.7k & Yes \\
        MuSiQue & 10.2k & 4.0k & Yes \\
        LongBench Pro & 1.5k--16k & 4.6k & No \\ 
        \bottomrule
    \end{tabular}
    \label{tab:datasets}
\end{table}

\paragraph{Baselines.}
We select two representative models as the instruction-following and reasoning base models - Qwen3-VL-8B-Instruct~\cite{yang2025qwen3} and GLM-4.1V-9B-Thinking~\cite{hong2025glm}.  
We employ \ours{} on both models to examine the effectiveness of VER heads for different tasks. 
To examine whether VER heads are critical to \ours{} framework, we adopt the following baselines:

\textbf{OCR-RAG}, which identifies recognition heads on the NIAH task following~\cite{baek2025large}; 

\textbf{Random-RAG}, which samples consecutive text segments of comparable length;

\textbf{Embedding RAG}, which utilizes the \textbf{Qwen3-VL-Embedding} model for retrieval;

\textbf{ColPali RAG}, which employs a VLM capable of producing high-quality multi-vector embeddings for retrieval;

\textbf{Glyph}, a specialized VLM trained for optimal visual context compression.

On each dataset, we report the score between ground truth and model prediction excluding thought tokens. We utilize the official evaluation scripts on all datasets, with F1-score acting as the primary metric. Within these, 40\% of the LongBench Pro data uses precision, and a part of HotpotQA uses exact matching.

\paragraph{Implementation.} 
\ours{} is a training-free approach that only has two hyper-parameters: (1) top-k VER heads; (2) top-N visual evidence patches. In experiments, we set $k=5$ to identify the top-5 VER heads that achieve the highest visual retrieval scores on development set $\mathcal{D}_\text{dev}$. In addition, we set $N=20$ to balance the amount of additional information against the compromised compression ratio. 
For RAG baselines including ours, we extract selected text by expanding the selected visual patches to rows. Then we append the additional context following the same evidence verbalization described in Equation~\ref{eq:verbalization}. Lastly, we provide the details of image rendering, VER head selection and prompt for all models in Appendix B.

%

\subsection{Performance (RQ1 and RQ2)}
We present the evaluation results of \ours{} and other baselines in Table~\ref{tab:ver-rag}. 
From the table, we have two key observations regarding the effectiveness (\textbf{RQ1}) and universality (\textbf{RQ2}) of the \ours{} framework.
\paragraph{1. Significant Improvement over Baselines.} Using Qwen3-VL-8B-Instruct as a comparative baseline, we observe that evidence retrieved via OCR heads yields only marginal performance gains. This validates our finding that low-level recognition heads lack the semantic alignment for reasoning—a gap \ours{} effectively bridges. Interestingly, random RAG can achieve comparable gains to OCR RAG, which supports the conclusion that OCR-related attention is rarely activated in long-context reasoning. Overall, GLM-4.1V-9B-Thinking+\ours{} almost achieves the best performance across all datasets. ColPali RAG shows a competitive performance on HotpotQA, however, this is attributed to the high dispersion of patches retrieved by ColPali, which improves recall at the cost of precision (Section~\ref{Retrieval Study}). This characteristic is particularly beneficial for the sparsely annotated Hotpot task.

\paragraph{2. Robustness across Model Architectures and Unseen Domains.}
As shown in the last column, \ours{} delivers over 20\% average relative gains across both model types and generalizes well to the unseen LongBench Pro. Notably, it boosts GLM-4.1V-9B-Thinking on DocMath by \textbf{85\%} (15.71$\to$29.02), confirming that reasoning models benefit from VER heads and validating the first high-entropy token as the ``retrieval moment''.

\subsection{Hyper-parameter Study (RQ3)}
\label{Ablation Study}
\begin{table}[t]
\centering
\small
\caption{F1 Scores under different values of $k$.}
\setlength{\tabcolsep}{4pt}
\begin{tabular}{lcccc}
\toprule
Dataset & 
$k = 5$ & 
$k = 10$ & 
$k = 15$ & 
$k = 20$ \\
\midrule
DocMath  & \textbf{9.45} & 8.60 & 9.39 & 9.35 \\
Qasper  & \textbf{36.37} & 35.91 & 36.10 & 35.45 \\
HotpotQA  & 32.32 & 33.36 &  32.88 &  \textbf{34.36} \\
MusiQue  & 17.84 & 18.01 &  18.00 &  \textbf{18.57} \\
\bottomrule
\end{tabular}

\label{tab:ablation-topk}
\end{table}
\paragraph{Hyperparameter Sensitivity.}
To explore the influence of the hyperparameter $k$ on the effect of \ours{}, we designed the hyperparameter experiment for exploration. We vary the value of $k$, and use the average distribution of the attention scores of $k$ VER heads to determine the retrieved patches. We compute the resulting F1 score to evaluate. The results are shown in Table~\ref{tab:ablation-topk}. We observe that larger $k$ values improve comprehension in long-context datasets (HotpotQA, MuSiQue), whereas $k=5$ is optimal for shorter contexts (DocMath, Qasper). However, the overall performance variance is not significant.

\begin{figure}
    \centering
    \includegraphics[width=1.0\columnwidth]{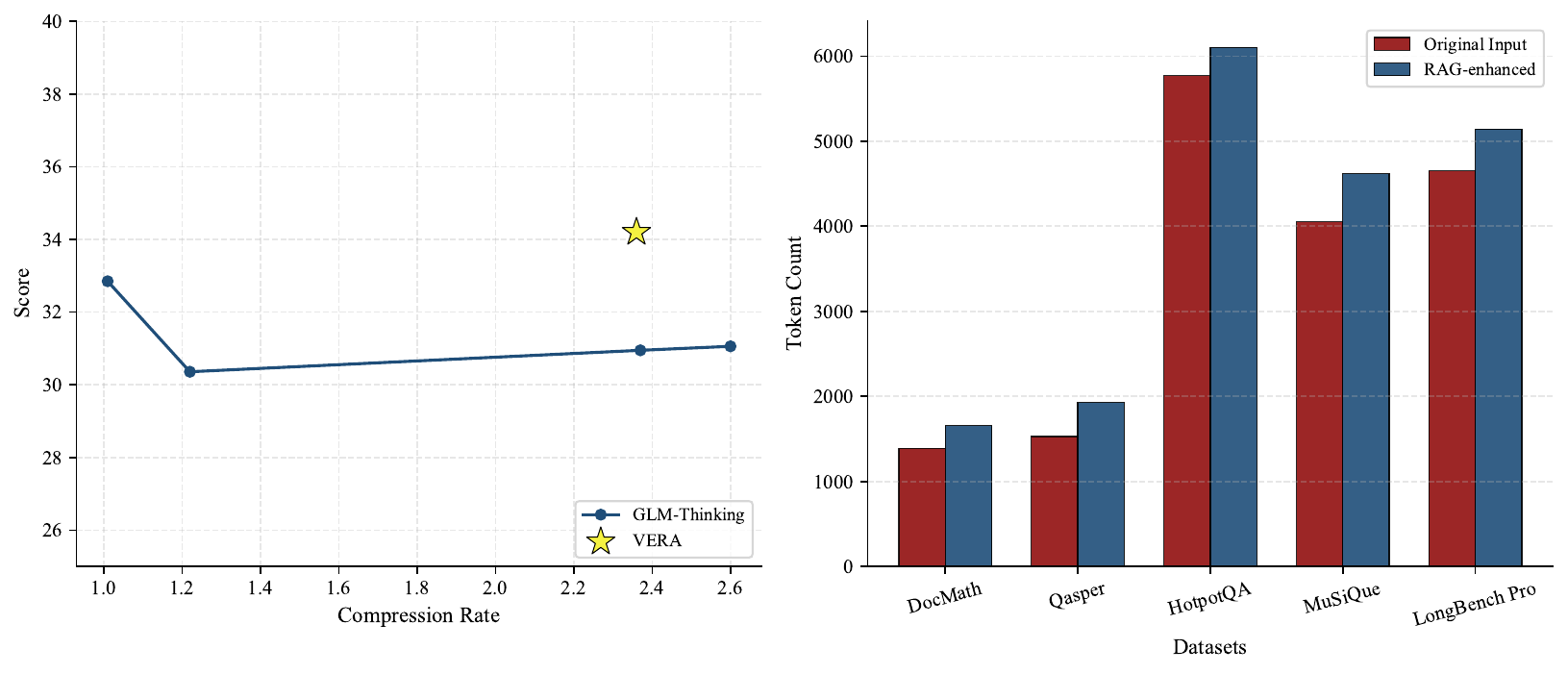} 
    \caption{Cost efficiency analysis: varying visual compression ratio (left) and the minimal token cost overhead of \ours{} (right).}
    \label{fig:cost_efficiency} %
\end{figure}

\paragraph{Cost Efficiency of \ours{}.}
Although \ours{} introduces additional context tokens to the input, we demonstrate that the base model (GLM-Thinking) consistently underperforms \ours{} across varying visual compression ratios in Figure~\ref{fig:cost_efficiency}(left). This indicates that the performance gains achieved via VER heads outweigh the minimal token overhead incurred by the additional textual context.
As shown in Figure~\ref{fig:cost_efficiency} (right), \ours{} achieves 20\% relative improvement with only 5.8\%--15\% additional tokens.

\subsection{Retrieval Performance Study (RQ4)}
\label{Retrieval Study}
\small
\begin{table*}[t] 
  \centering
  \caption{Retrieval performance of Qwen3-VL-8B Embedding, ColPali and VER heads.}
  \resizebox{\textwidth}{!}{
  \begin{tabular}{lcccccccccccc}
    \toprule
    \multirow{2}{*}{Model} & \multicolumn{3}{c}{Qasper} & \multicolumn{3}{c}{DocMath} & \multicolumn{3}{c}{Musique} & \multicolumn{3}{c}{Hotpot} \\

    \cmidrule(lr){2-4} \cmidrule(lr){5-7} \cmidrule(lr){8-10} \cmidrule(lr){11-13}

     & Prec. & Rec. & F1 & Prec. & Rec. & F1 & Prec. & Rec. & F1 & Prec. & Rec. & F1 \\

    \midrule

    Qwen3-VL-embedding & 0.3679 & 0.4354 & 0.3510 & 0.1282 & 0.2755 & 0.1575 & 0.1609 & 0.5023 & 0.2336 & 0.0621 & 0.5012 & 0.1063 \\

    ColPali & 0.2913 & \underline{0.5454} & 0.3365 & \underline{0.2799} & \underline{0.4893} & \underline{0.3112} & 0.2110 & \textbf{0.6409} & \underline{0.3018} & 0.0779 & \textbf{0.7183} & 0.1362 \\

    VER Heads (confidence) & \underline{0.3731} & 0.5435 & \underline{0.3919} & 0.2527 & 0.2975 & 0.2276 & \underline{0.2146} & 0.5718 & 0.2967 & \underline{0.0933} & 0.6172 & \underline{0.1544} \\

    VER Heads (ours) & \textbf{0.4142} & \textbf{0.5629} & \textbf{0.4223} & \textbf{0.4418} & \textbf{0.5404} & \textbf{0.4214} & \textbf{0.2393} & \underline{0.5948} & \textbf{0.3251} & \textbf{0.1065} & \underline{0.6815} & \textbf{0.1753} \\

    \bottomrule

  \end{tabular}
  }


\end{table*}
Having demonstrated the effectiveness and universality of \ours{} on QA tasks, it is essential to further verify its capabilities through retrieval experiments. To this end, we conducted a comparative study of the retrieval performance of \ours{} against other strong baselines.

\paragraph{Baselines.}As described in Section \ref{setup}, we utilize the top 20 VER Heads identified on $\mathcal{D}_\text{dev}$, which correspond to the heads with the highest average VER scores. Observing that heads with higher confidence (\ie{low attention entropy}) typically yield higher VER scores, we establish a baseline using the retrieval results from the 5 most confident heads within this top 20 set. Furthermore, we employ the end-to-end retrieval model ColPali and the Qwen3-VL-embedding model as additional baselines for comparison. Considering variations in patch size, we fixed the number of patches retrieved by Colpali at 20. This ensures consistency in the number of tokens used for text augmentation across all baselines.

\paragraph{Results.}Experimental results demonstrate that the VER heads in \ours{} achieve the highest F1 scores across all datasets, which is highly consistent with their performance on QA tasks. Notably, even ColPali—a Vision Language Model trained to produce high-quality multi-vector embeddings from document images—only outperforms our method on a subset of Recall metrics, a gain achieved at the expense of Precision. On the Qasper dataset, our F1 score surpasses Colpali by 8.58\%.

\section{Conclusions}

In this work, we identify \textbf{Visual Evidence Retrieval (VER) Heads}—a sparse, dynamic set of attention heads distinct from static OCR heads that is causal to long-context visual reasoning. Through systematic analysis, we demonstrate that VER heads are (1) universal across datasets (Spearman correlation $>0.44$), (2) functionally critical (masking them causes performance degradation) and (3) enhanced through training.

Building on this insight, we propose \ours{}, a training-free framework that detects model uncertainty via entropy spikes and explicitly verbalizes the visual evidence attended by VER heads. \ours{} requires no parameter updates and achieves average relative improvements of 21.3\% on Qwen3-VL-8B-Instruct and 20.1\% on GLM-4.1V-Thinking across five benchmarks. Our work establishes a new paradigm for interpretable VLM enhancement by transforming mechanistic understanding into actionable inference-time interventions. Future work may explore zero-shot VER head identification and adaptive visual-text compression guided by attention dynamics.



\newpage

\bibliographystyle{named}
\bibliography{ijcai26}

\clearpage

\section*{Appendix}

\section{Implementation Details}

\subsection{\ours{} Algorithm}

The Algorithm~\ref{alg:entropy_rag} shows our detailed \ours{} logic. We designed the algorithm to utilize KV Cache efficiently, enable memory-efficient attention extraction and accelerate result generation. 
By reducing GPU memory requirements during attention computation, the method allows the experiments to be executed on a single NVIDIA A100 80G GPU.

\begin{algorithm}[h]
\SetAlgoLined
\KwIn{Context $C$, Question $Q$, Images $I$}
\KwOut{Generated Answer $A$}

Initialize prompt $P \leftarrow [I, C, Q]$\;

$KV \leftarrow \mathcal{M}_{\text{fast}}.\text{prefill}(P)$\;

$y_{0...t-1} \leftarrow \emptyset$, $Triggered \leftarrow \text{False}$\;

\For{$t \leftarrow 1$ \KwTo $T_{\max}$}{
    $L_t, KV \leftarrow \mathcal{M}_{\text{fast}}.\text{decode}(y_{t-1}, KV)$\;
    $H_t \leftarrow \text{Entropy}(\text{Softmax}(L_t))$\;
    
    \If{$\text{IsHighEntropy}(H_t)$ \textbf{and not} $Triggered$}{
        $Triggered \leftarrow \text{True}$\;
        \tcp{Compute attention using monitor model and saved cache}
        $A_t \leftarrow \mathcal{M}_{\text{attn}}.\text{forward}(P, y_{0...t-1}, KV)$\;
        
        $S_{attn} \leftarrow \frac{1}{20} \sum_{h=1}^{20} A_t^{(h)}$\;
        
        $E \leftarrow \text{RetrieveEvidence}(I, S_{attn})$\;
        $P_{new} \leftarrow [I, C, Q, E]$\;
        \Return $\text{VLM}(P_{new})$\;
    }
    
    $y_t \leftarrow \operatorname{argmax}(L_t)$\;
    Append $y_t$ to $y_{0...t-1}$\;
}
\Return $\text{Detokenize}(y_{0...T})$\;

\caption{Entropy-Triggered \ours{}}
\label{alg:entropy_rag}
\end{algorithm}


\subsection{Render Text as Image}

Our image rendering tool leverages the open-source code in Glyph. First, the rendering environment is initialized with configuration parameters, like font path, size, color, and page dimensions (The detailed config file is shown in \ref{fig:render config}). Next, the ReportLab library is employed to convert input text into PDF format according to specified typesetting styles. Subsequently, the pdf2image library converts the generated PDF documents into high-resolution PNG images. These images undergo post-processing, including horizontal scaling, adaptive width cropping, and the vertical concatenation of multi-page images. Additionally, the algorithm features an evidence text highlighting function, which renders specific evidence sentences in red to facilitate data annotation.
\begin{figure}
\small
\centering
\begin{tcolorbox}[
  enhanced,
  title=Rendering Config,
  separator sign={\tcbline},
  separator sign dash={3pt}{3pt},
]
\{

      "page-size": "595,842",
      
      "dpi": 72,
      
      "margin-x": 10,
      
      "margin-y": 10,
      
      "font-path": "config/Verdana.ttf",
      
      "font-size": 9,
      
      "line-height": 10,
      
      "font-color": "\#000000",
      
      "alignment": "LEFT",
      
      "horizontal-scale": 1.0,
      
      "first-line-indent": 0,
      
      "left-indent": 0,
      
      "right-indent": 0,
      
      "space-after": 0,
      
      "space-before": 0,
      
      "border-width": 0,
      
      "border-padding": 0,
      
      "page-bg-color": "\#FFFFFF",
      
      "para-bg-color": "\#FFFFFF",
      
      "auto-crop-width": true,
      
      "auto-crop-last-page": true
      
\}
\end{tcolorbox}
\vspace{-2mm}
\caption{Our Rendering Config in main experiments.}
\label{fig:render config}
\end{figure}

\subsection{Prompts}

Figure~\ref{fig:raw prompt}, Figure~\ref{fig:rag prompt} shows the detailed prompts we used in experiments.

\begin{figure}
\small
\centering
\begin{tcolorbox}[
  enhanced,
  title=Original prompts for Qwen3-VL-8B-Instruct and GLM-4.1V-9B-Thinking.,
  separator sign={\tcbline},
  separator sign dash={3pt}{3pt},
]
Please answer the question based on the document images provided.

\{question\}

Please output your answer **directly** based on the provided images and text.
\end{tcolorbox}
\vspace{-2mm}
\caption{Prompts for original answer generation of Qwen3-VL-8B-Instruct and GLM-4.1V-9B-Thinking.}
 
\label{fig:raw prompt}
\end{figure}

\begin{figure}
\small
\centering
\begin{tcolorbox}[
  enhanced,
  title=RAG prompts for Qwen3-VL-8B-Instruct and GLM-4.1V-9B-Thinking.,
  separator sign={\tcbline},
  separator sign dash={3pt}{3pt},
]
Please answer the question based on the document images provided.

\{question\}

Some text Information (Maybe useful, extracted from image): \{rag\_info\} .

Judge whether you need it or not first, **do not** hesitate repeatedly. The answer shouldn't include reason (if not required).

Please output your answer **directly** based on the provided images and text.
\end{tcolorbox}
\vspace{-2mm}
\caption{Prompts for RAG answer generation of Qwen3-VL-8B-Instruct and GLM-4.1V-9B-Thinking.}
 
\label{fig:rag prompt}
\end{figure}

\section{Experiment Details}
\subsection{Methods for Dataset Enhancement and Clean}
We implemented dataset-specific augmentation strategies focused on context expansion. 
For the HotpotQA dataset, we merge the contexts of ten QA pairs into a single unified context, such that every ten questions share the same context. For the MuSiQue dataset, we apply a similar strategy by fusing six contexts, where six QA pairs share one common context. As a result, the effective input length of both HotpotQA and MuSiQue is extended to over 10K tokens. This data augmentation strategy is inspired by the context concatenation method used in LongBench.

In addition, we cleaned the LongBenchPro dataset by first filtering out all Chinese QA pairs. Then we selected samples with context lengths between 1K and 16K tokens, which were used to build our evaluation dataset. This is because we empirically observe that 8B-scale models are more likely to produce repetitive outputs when handling longer contexts, thereby compromising the reliability of the experimental results.

\subsection{Detailed Visual Retrieval Score and OCR Score Distribution Heatmap}
\label{sec:Detailed-distribution}
\begin{figure*}[t]
    \centering

    \begin{subfigure}[b]{0.32\textwidth}
        \centering
        \includegraphics[width=\textwidth]{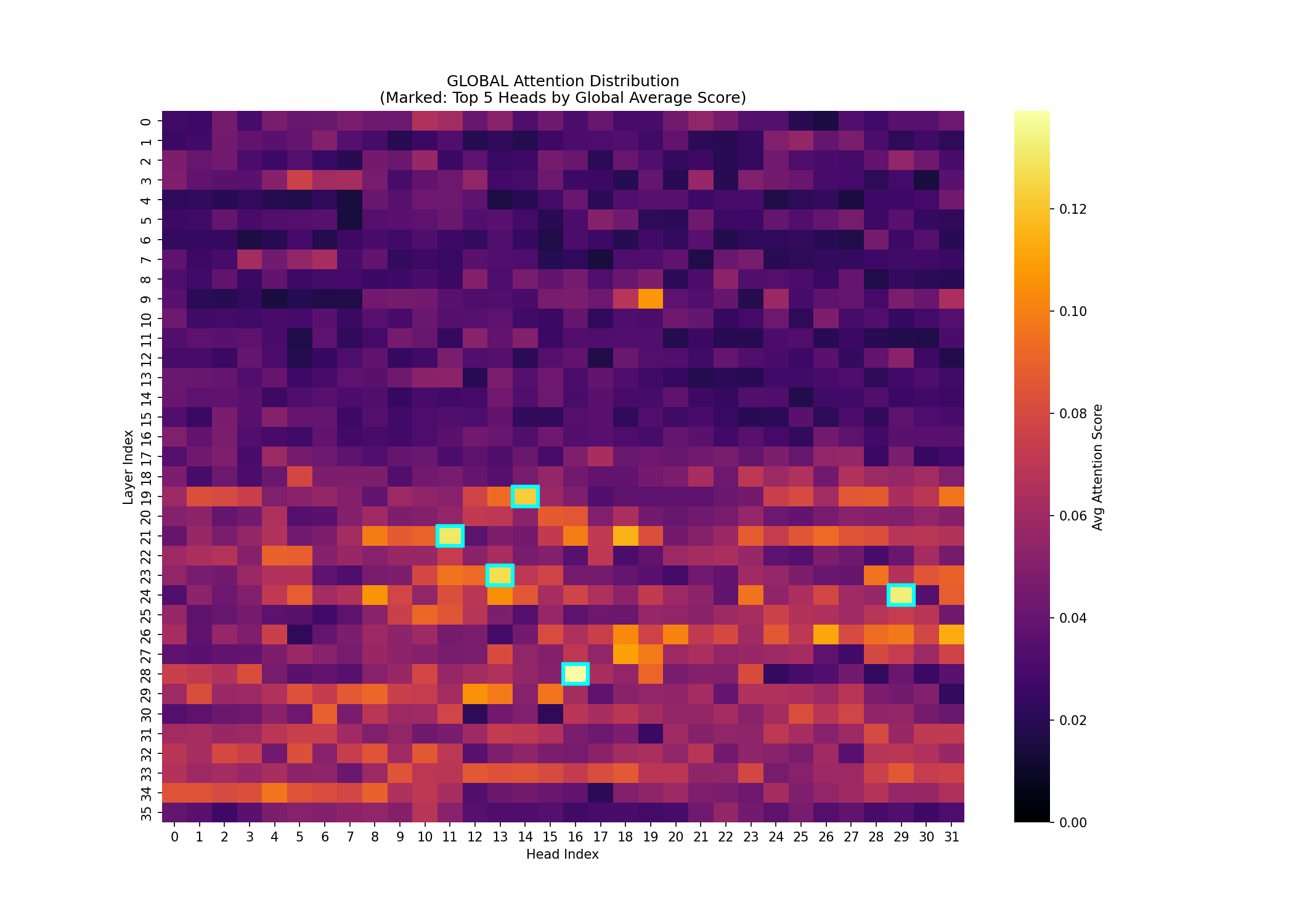}
        \caption{DocMath}
    \end{subfigure}
    \hfill 
    \begin{subfigure}[b]{0.32\textwidth}
        \centering
        \includegraphics[width=\textwidth]{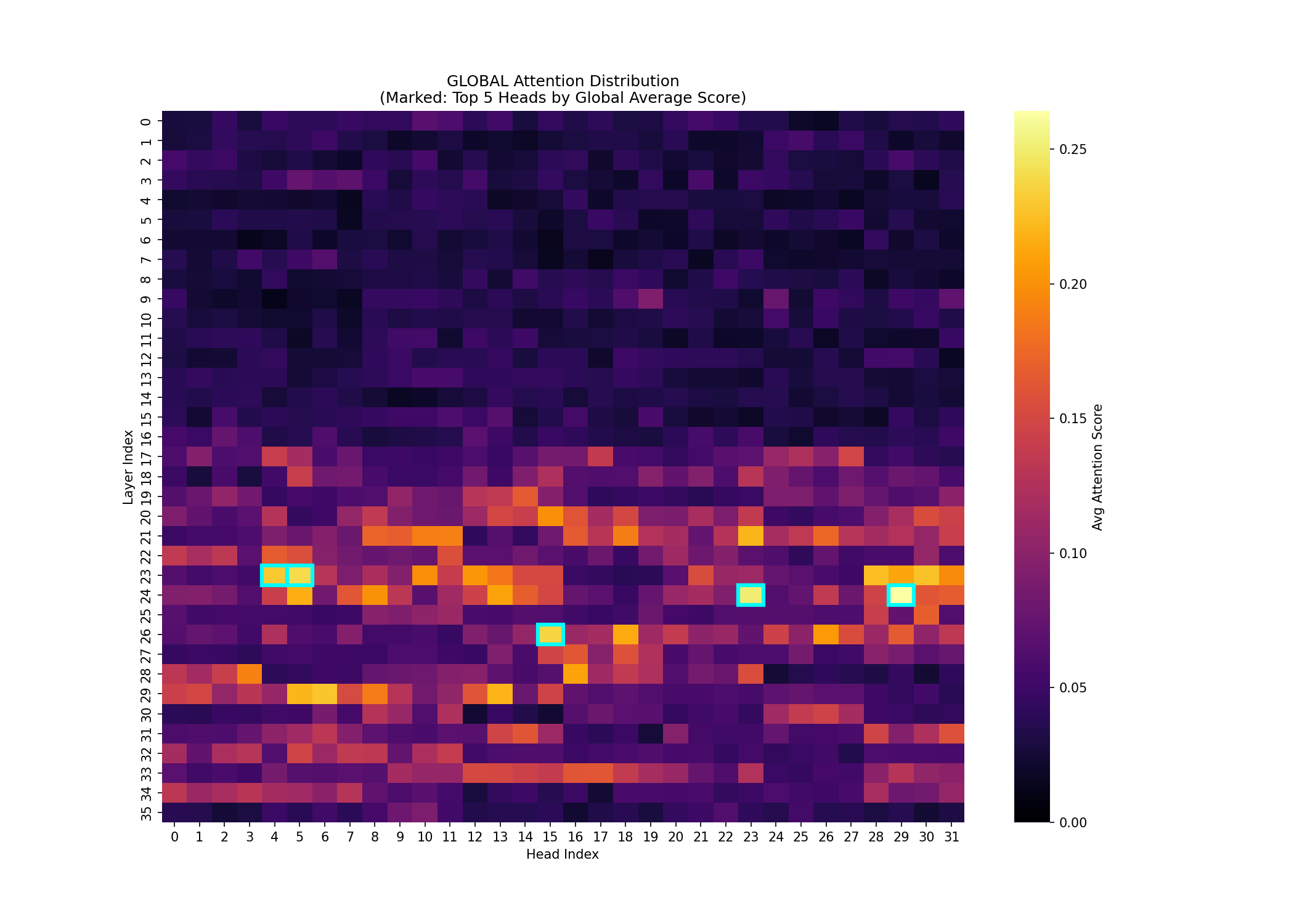}
        \caption{DocMath (SFT)}
    \end{subfigure}
    \hfill
    \begin{subfigure}[b]{0.32\textwidth}
        \centering
        \includegraphics[width=\textwidth]{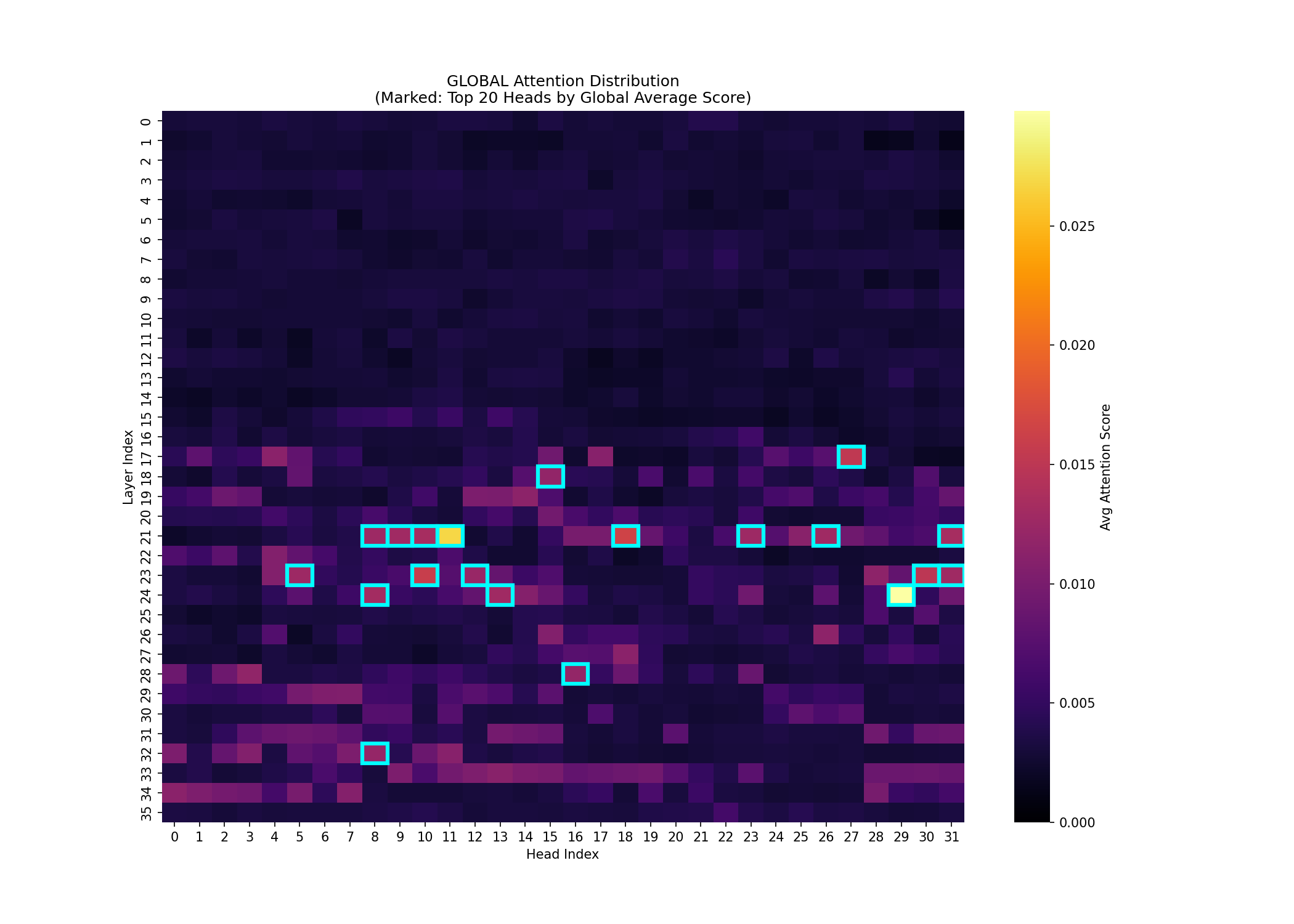}
        \caption{HotpotQA}
    \end{subfigure}

    \vspace{0.5cm}

    \begin{subfigure}[b]{0.32\textwidth}
        \centering
        \includegraphics[width=\textwidth]{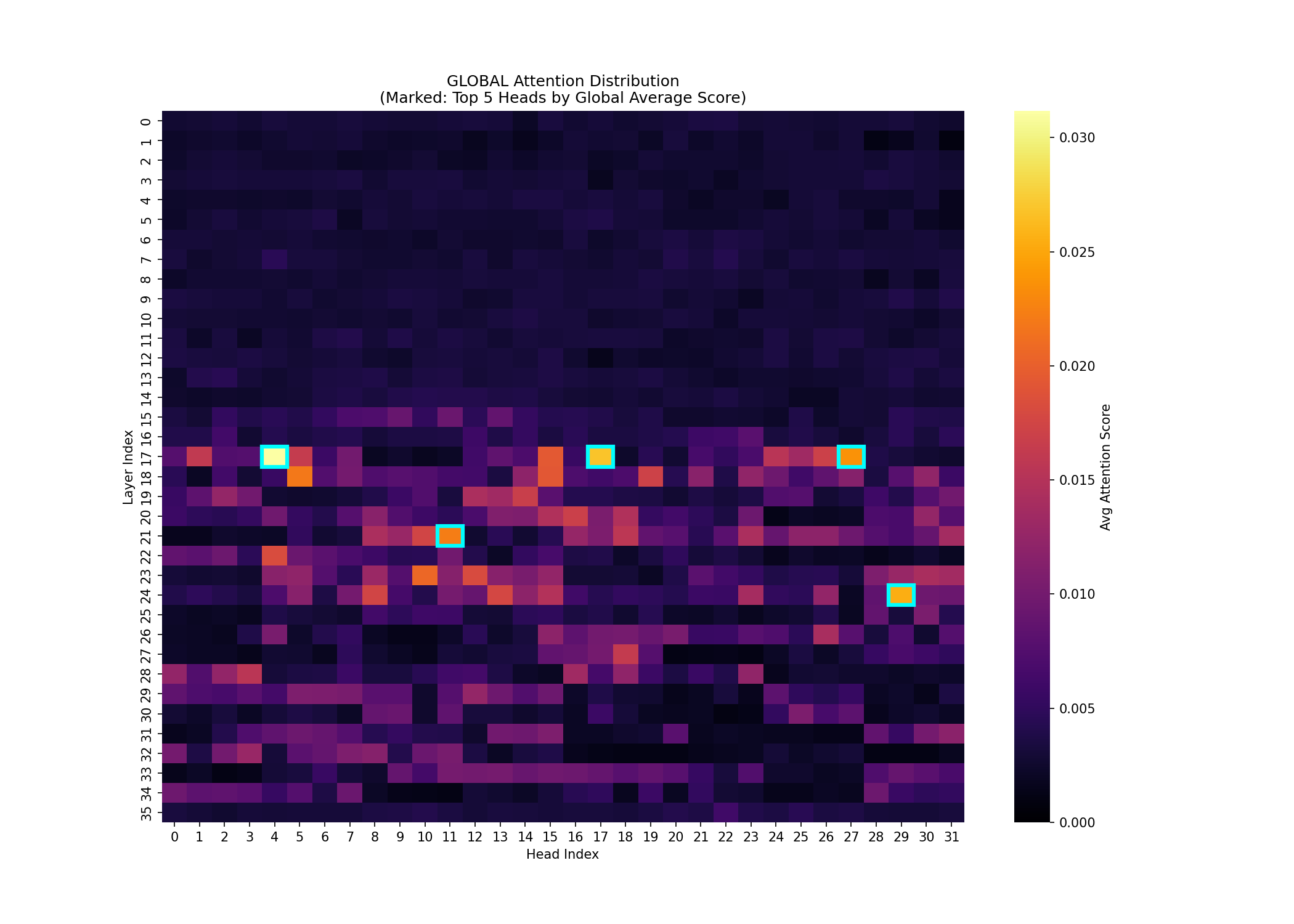}
        \caption{HotpotQA (SFT)}
    \end{subfigure}
    \hfill
    \begin{subfigure}[b]{0.32\textwidth}
        \centering
        \includegraphics[width=\textwidth]{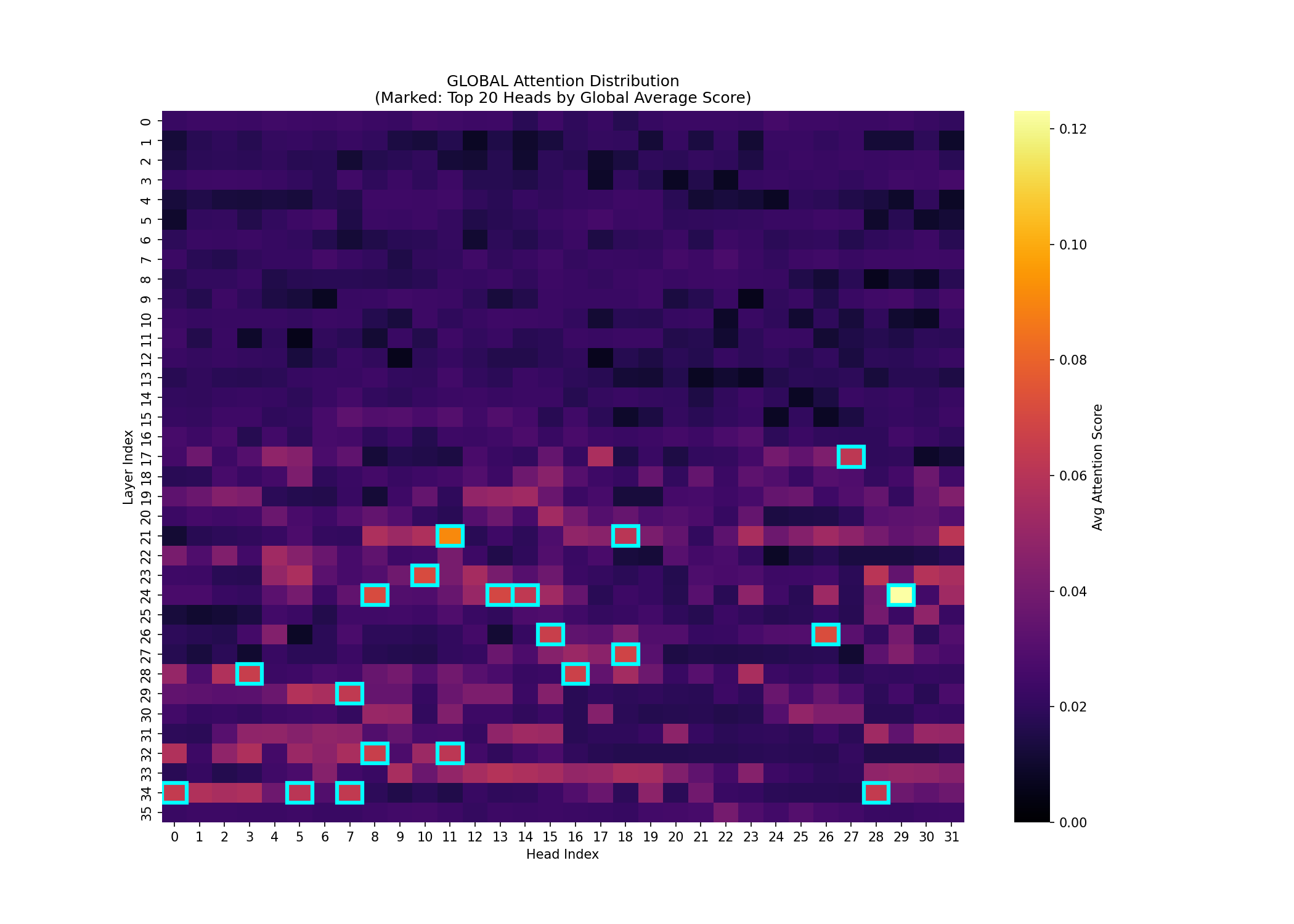}
        \caption{Musique}
    \end{subfigure}
    \hfill
    \begin{subfigure}[b]{0.32\textwidth}
        \centering
        \includegraphics[width=\textwidth]{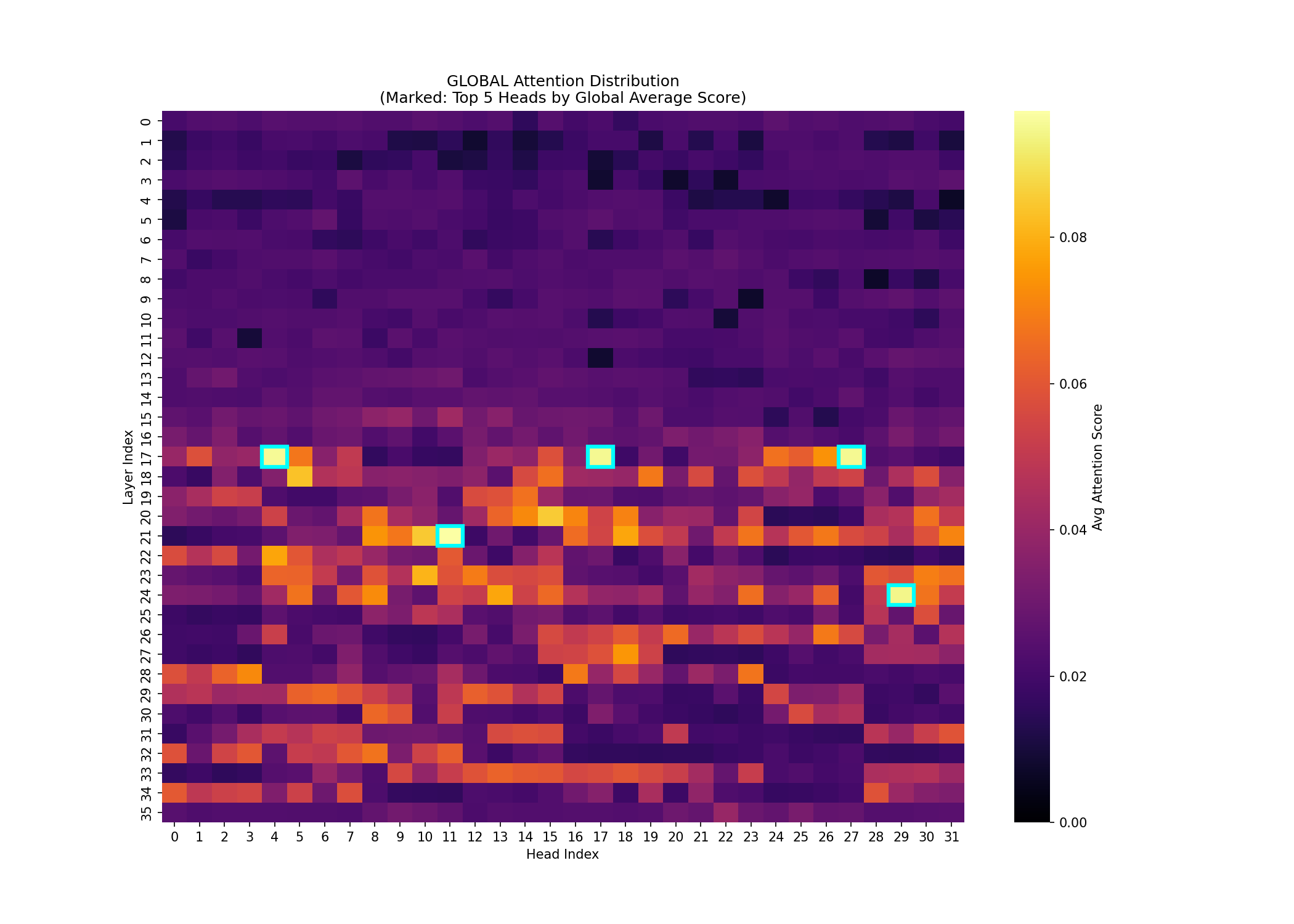}
        \caption{Musique (SFT)}
    \end{subfigure}

    \vspace{0.5cm}

    \begin{subfigure}[b]{0.32\textwidth}
        \centering
        \includegraphics[width=\textwidth]{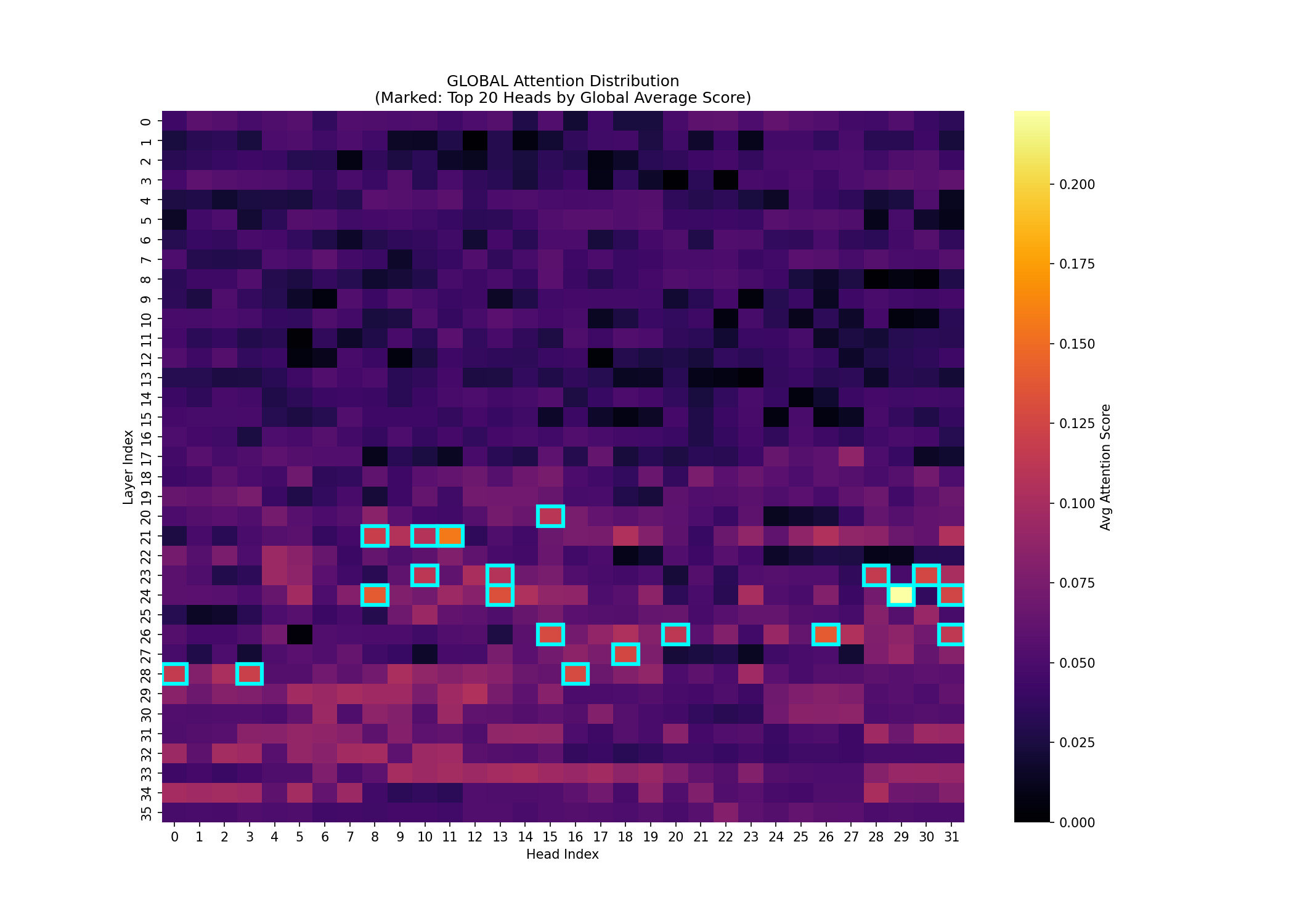}
        \caption{Qasper}
    \end{subfigure}
    \hfill
    \begin{subfigure}[b]{0.32\textwidth}
        \centering
        \includegraphics[width=\textwidth]{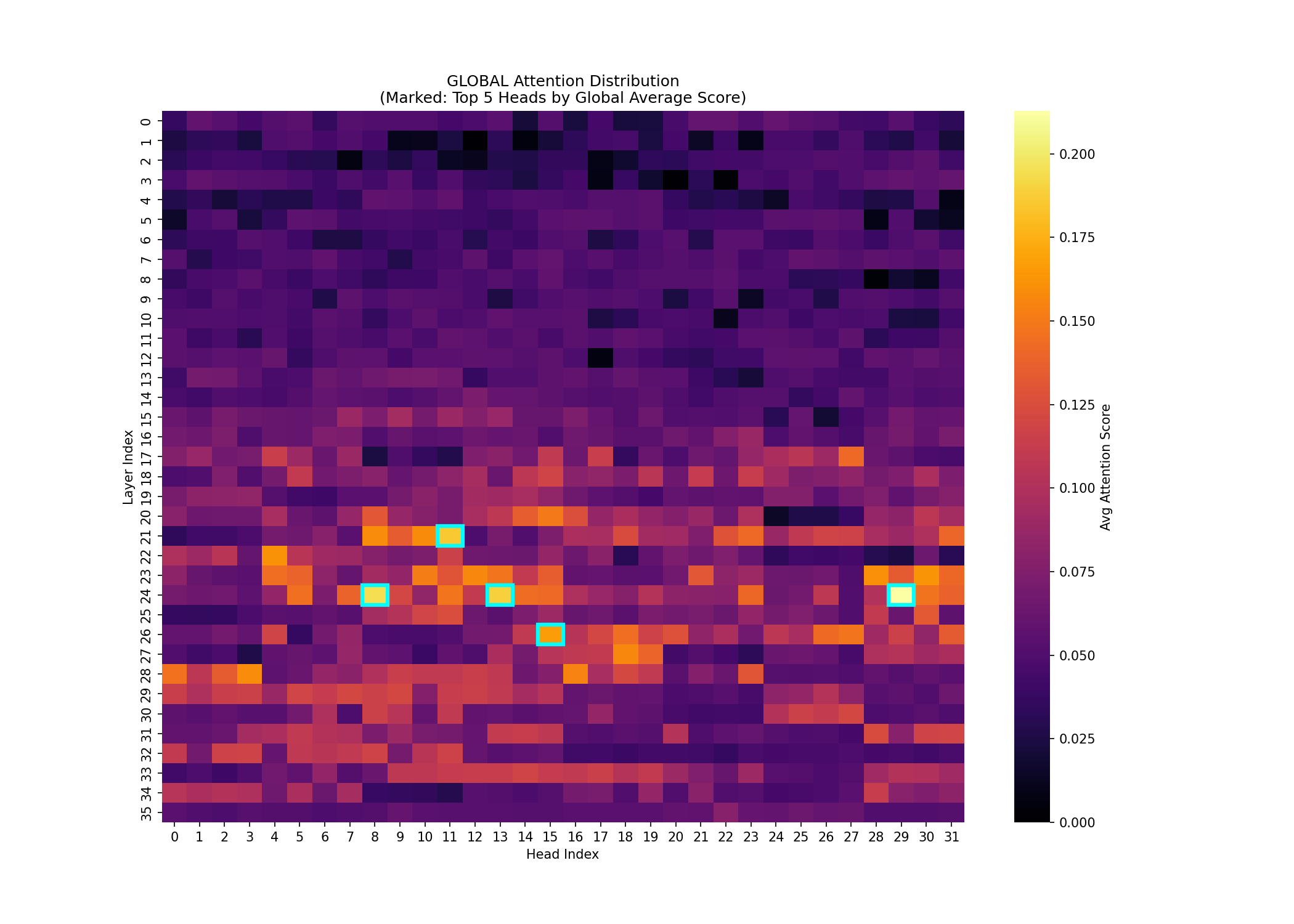}
        \caption{Qasper (SFT)}
    \end{subfigure}
    \hfill
    \begin{subfigure}[b]{0.32\textwidth}
        \centering
        \includegraphics[width=\textwidth]{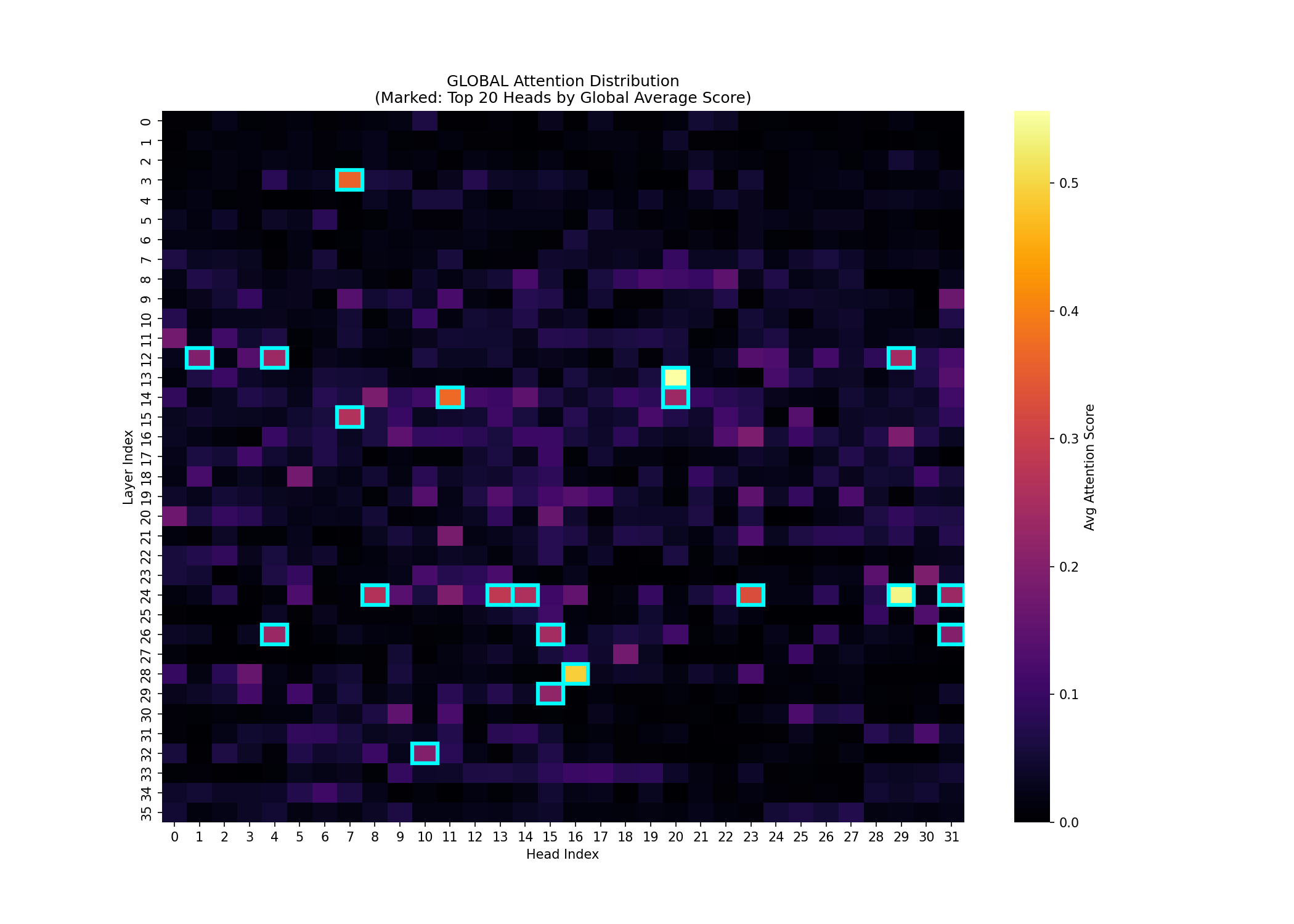}
        \caption{NIAH OCR Score} 
    \end{subfigure}

    \caption{Combined visualizations across datasets. (a)-(h) show Visual Retrieval Scores before and after supervised fine-tuning (SFT). (i) shows OCR Scores on the NIAH dataset.}
    \label{fig:heatmap_3x3_all}
\end{figure*}

The Figure~\ref{fig:heatmap_3x3_all} illustrates the distribution of visual retrieval scores of the Qwen3-VL-8B-Instruct model across four different datasets and the distribution of OCR scores of the same model on the NIAH dataset, where the evaluation protocol is consistent with that used in the original OCR head paper. The figure shows that the distributions of VER scores are similar across various datasets and are different from OCR score distribution.

\end{document}